%% file: iclr2027_conference.tex
\documentclass{article}

\usepackage{iclr2027_conference,times}
\iclrfinalcopy

\input{math_commands.tex}

\usepackage{graphicx}
\usepackage{amsmath}
\usepackage{amssymb}
\usepackage{booktabs}
\usepackage{xspace}
\usepackage{xcolor}
\usepackage{subcaption}
\usepackage{tikz}
\usepackage{enumitem}
\usetikzlibrary{arrows.meta,positioning,calc}
\usepackage[export]{adjustbox}   %

\usepackage{hyperref}
\usepackage{url}

\usepackage[capitalize]{cleveref}
\crefname{section}{Sec.}{Secs.}
\Crefname{section}{Section}{Sections}
\Crefname{table}{Table}{Tables}
\crefname{table}{Tab.}{Tabs.}

\input{commands}

\usepackage{float}
\title{Just Image Transformers with Late-Stage Pixel Refinement}

\author{
Denis Lukovnikov \\
Ruhr University Bochum \\
\texttt{denis.lukovnikov@rub.de}
\And
Andreas Müller \\
Ruhr University Bochum \\
\texttt{andreas.mueller-t1x@rub.de}
\And
Asja Fischer \\
Ruhr University Bochum \\
\texttt{asja.fischer@rub.de}
}

\begin{document}

\maketitle
\lhead{Originally submitted to ICLR 2027}

\begin{abstract}
Recent work on latent-free diffusion transformers has demonstrated the feasibility of training image generators without a latent VAE.
This avoids compression artifacts from the VAE and simplifies the training
pipeline.
In particular, JiT showed that even a standard transformer can be trained to
generate images when the model is trained to predict the final image
($x$-prediction).
Even though JiT unlocks training efficiency comparable to latent models, it
still suffers from reduced image quality likely arising from modeling the image as
a collection of large patches.
In this work, we revisit JiT and show that adding simple convolutional pixel-pathway layers in
a post-training phase significantly improves generation quality.
Moreover, we show that the added pixel-pathway does not need to be active in the early denoising stages, and can be activated as late as in the last $20\%$ of the denoising process.
Our experiments on ImageNet@256 show improvements compared to the original JiT results, achieving an FID of $1.78$, without using additional losses.
\end{abstract}

\section{Introduction}
\label{sec:intro}

Recent advances~\citep{jit,pixeldit,dip,pixelgen,l2p,pixelu,pixsgr} in latent-free generation avoid the bottleneck of latent-based methods by training diffusion models directly in pixel space.
Such pixel-space diffusion transformers (pDiTs) take an RGB image as input and predict a denoising step in pixel space.
\cite{jit} observe that the v-prediction objective commonly used to train diffusion transformers requires the network to predict off-manifold targets because of the added noise component.
They instead propose predicting the final image (\textit{x-prediction}), thereby keeping the prediction targets on the manifold of real images.
Their experiments show that, where v-prediction fails with large patches, x-prediction succeeds with patches of up to $32\times32$ pixels.
Their model is a standard image transformer that divides an image into large, non-overlapping patches, hence the name Just Image Transformer (JiT).

Although these results show that simple large-patch transformers can be trained directly in pixel space, we find that JiT outputs lack sharpness and show subtle patch-boundary artifacts (see \cref{fig:blurriness,fig:artifacts}).
We hypothesize that these issues mainly arise from the architectural choices.
In fact, JiT independently decodes final patch tokens into non-overlapping pixel patches.
This requires consistency across patch boundaries to be established through global patch-level attention.
Moreover, the output head has no direct access to the noisy input $\x_t$, so recoverable fine detail must pass through the input bottleneck and the entire transformer backbone.
These architectural choices place the burden of resolving local dependencies, such as edge continuation and texture alignment, on a large-patch transformer, which is inefficient use of global attention and transformer capacity.
This motivates using dedicated local processing that communicates across patch boundaries.

In this work, we study the architectural and training requirements for pixel-space diffusion transformers to learn fine, spatially coherent details.
We augment the patch-level transformer with a pixel-level pathway and study its late activation. We also revisit the timestep-sampling and loss-weighting strategies used during training.
Our modifications can be added to pretrained JiT checkpoints through fine-tuning, avoiding the cost of training the full model from scratch.

Our contributions are as follows:
\begin{enumerate}[leftmargin=*,label=\arabic*.,labelsep=0.5em,itemsep=0.35em,topsep=0.5em]
    \item \textbf{Post-training refinement.}
    We improve JiT models by adding a pixel-level pathway consisting of only a few convolutional blocks on top of the big-patch transformer.
    Unlike previous work, we do so in a post-training manner, showing it's possible to pre-train just a transformer backbone using an $x$-prediction objective and add pixel refinement in a shorter fine-tuning stage.

    \item \textbf{Efficient late-stage processing.}
    We show that, when combined with $x$-prediction, the pixel-level pathway is not necessary throughout the entire denoising process.
    Consequently, we propose to use the pixel pathway only in the last $\leq25\%-50\%$ of denoising steps and show this has minimal to no impact on image quality compared to full-time application.
    This result further demonstrates the promise of $x$-prediction for pixel-space modeling while reducing generation cost compared to previously proposed methods with a dedicated pixel pathway.

    \item \textbf{Thorough analysis.} We revisit the training protocol, showing that the timestep distribution during training has an effect on learning fine details and the timesteps near the clean endpoint benefit from better coverage. We also experiment with different head configurations and revisit the generation procedure of JiT, proposing a correction for the final denoising steps.
\end{enumerate}

\begin{figure}[t]
  \centering
  \begin{minipage}[t]{0.554\linewidth}
    \centering
    \resizebox{\linewidth}{!}{%
      \input{figures/blurriness/blurriness}%
    }
    \captionof{figure}{Loss of sharpness in original JiT (left) vs. adding our pixel pathway (right). Measured sharpness (Laplacian variance) reported above.}
    \label{fig:blurriness}
  \end{minipage}
  \hfill
  \begin{minipage}[t]{0.405\linewidth}
    \centering
    \resizebox{\linewidth}{!}{%
      \input{figures/artifacts/artifacts}%
    }
    \captionof{figure}{Patch boundaries in JiT-L. The right image is sharpened to accentuate the patch borders.}
    \label{fig:artifacts}
  \end{minipage}

\end{figure}

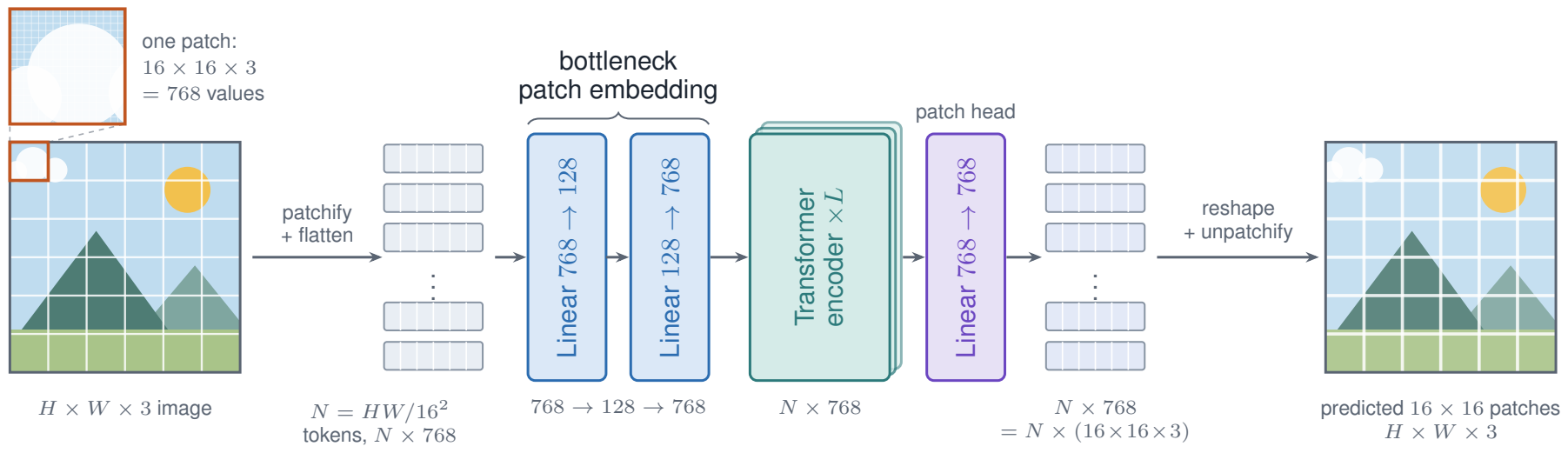
\begin{figure}[t]
  \centering
  \adjustbox{trim={1cm} 0 0 0, clip, width=0.85\linewidth}{\input{figures/jit_diagram}}
  \caption{The architecture used by \cite{jit} is an image transformer with an input bottleneck.}\label{fig:jit}
\end{figure}

\section{Approach}
\label{sec:approach}
In this work, we investigate the addition of a \textbf{small pixel-level prediction head} that enables communication across the patch boundaries of JiT's original transformer.
Note that while the general idea of using a dedicated pixel-level pathway has been explored in recent work, some of the existing methods do not enable information sharing across big patches in the pixel-level pathway (DiP~\citep{dip}), or still use global attention over recombined big-patches (PixelDiT~\citep{pixeldit}).

In addition to model modification, we propose a \textbf{two-stage training process}. First, the large-patch backbone is pre-trained as described by~\cite{jit}.
Then, in an additional training phase, the model is adapted and further fine-tuned with an emphasis on recovering high-frequency details in images that plain JiT wasn't able to learn.
We believe this enables more efficient training, since JiT training is the most efficient among recent pDiT's~\citep{l2pzimage} as it doesn't require running an additional compute-heavy pixel head throughout the entire training.

Below, we first give an overview of JiT (\cref{sec:preliminaries_jit}) and discuss the simple convolution pixel head that operates on small $2\times 2$ patches (\cref{sec:convhead}).
Then, we propose a more efficient generation procedure that saves compute by running the pixel head only in later denoising stages (\cref{sec:partialhead}).

\subsection{Preliminaries: JiT}
\label{sec:preliminaries_jit}
\cite{jit} use a standard transformer architecture with a bottleneck at the input.
Given an RGB image $x \in \R^{H \times W \times 3}$, it is first rearranged into a collection of large patches of size $s_p$ (e.g. patches of $16\times 16$ pixels): $p \in \R^{h \times w \times (3 * s_p^2)}$, where $h = H / s_p$ and $w = W / s_p$.
Each patch is then transformed using a bottleneck layer to a patch token.
The patch tokens are then processed by a transformer and the output image is produced simply by mapping updated patch features back to pixel space using a linear layer.
This transformer has a similar size and computational complexity as its latent counterparts.
JiT's model is illustrated in \cref{fig:jit}.

\cite{jit} show that when the model is made to predict $v=x - \epsilon$ (v-prediction for rectified flows), training fails with large patch sizes.
This is explained by the fact that %
the velocity target $v$ contains a noise component and does not lie on the hypothesized low-dimensional manifold of natural images. %
However, they found that requiring the model predict $x$ (x-prediction) instead enables the use of large patches (up to $s_p = 64$ for $1024\times 1024$ images was tested).
They convert $x$-predictions to velocities,
$\vpred = (\xpred - \x_t)/(1-t)$, and minimize
$
\mathcal{L}_{v}
=
\mathbb{E}_{x,\epsilon,t}
[
\left\|\vpred-v\right\|_2^2
]
$.

\subsection{Convolutional Pixel Head}
\label{sec:convhead}
We replace the original linear output layer of the JiT transformer with a stack of ConvNeXt-inspired ~\citep{convnext} blocks.
Given the noisy image $\x_t$, it first passes through the JiT big-patch transformer resulting in a hidden representation
\begin{align}
    h &= b(\x_t, t, c) \enspace ,
\end{align}
where $c$ is the condition embedding.
Subsequently, the pixel head is used to obtain a clean-image estimate,
\begin{align}
    \xpred &= g(h, \x_t, t, c) \enspace . \label{eq:pixelhead}
\end{align}
Note that the noisy image $\x_t$ is also passed directly to the pixel head, along with the backbone features $h$.

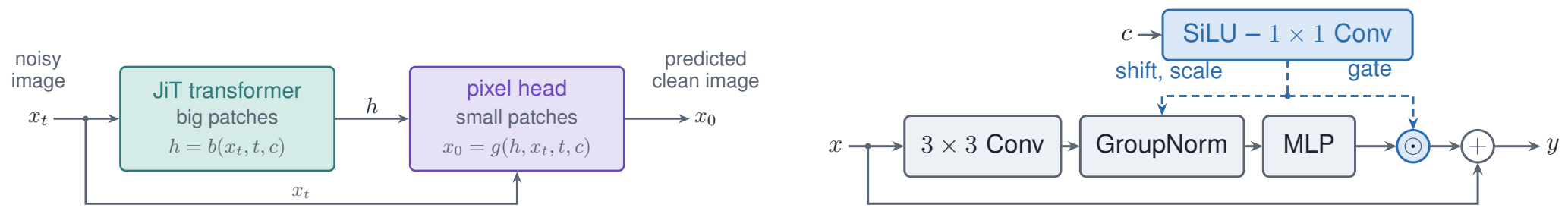
\begin{figure}[t]
  \centering
  \begin{minipage}[b]{0.5\linewidth}
    \centering
    \resizebox{\linewidth}{!}{\input{figures/convjit_diagram}}
  \end{minipage}%
  \hfill
  \begin{minipage}[b]{0.49\linewidth}
    \centering
    \resizebox{\linewidth}{!}{\input{figures/block_diagram}}
  \end{minipage}
  \caption{Our model architecture (left) and ConvNeXt-inspired pixel-head block architecture (right). Note the similarity of the ConvNeXt block to a transformer block.}
  \label{fig:jitconvdiagram}
\end{figure}

The pixel head consists of a stack of residual ConvNeXt-inspired blocks (see \cref{fig:jitconvdiagram}), each containing a dense
$3\times3$ convolution and two $1\times1$ convolutions.
The noisy image $\x_t$ is first split into small non-overlapping patches ($2\times 2$), and embedded before being fed to the stack of convolutional blocks.
We use small patches (rather than pure pixels like previous work) for increased computational efficiency.
Note that unlike previous work that used convolutions (DiP's UNet), our pixel head operates over the entire image and is not constrained to a single big patch.
This enables low-level communication across large-patch boundaries.%

\paragraph{Why convolutions?}
In early experiments, we explored more advanced architectures for the pixel head like using sliding window attention with small windows.
Both convolutions and sliding window attention implement the inductive bias that local texture matching and patch boundary continuity should be ensured only using local low-level features.
While sliding window attention was effective as well, we found it to be significantly more expensive and thus omitted it from further experiments.

\paragraph{Conditioning on backbone features.}
To condition the pixel head on the backbone's features $h$, they are first linearly projected to a set of small-patch conditioning vectors $c_b^{x_p, y_p}$, one for each small $2\times2$ patch (indexed by $(x_p, y_p)$) contained in the big $16\times16$ backbone patch.
These are then simply added with the small-patch embeddings of $\x_t$ before feeding them to the convolutional stack.
While this simple approach appears to work well, the alternative is to pass each small-patch vector $c_b^{x_p, y_p}$ as conditioning to each convolutional block in the stack, where it is injected using AdaLN modulation. We tested this as well but found it to perform similarly and thus opted for the simpler additive version.

\subsection{Late-stage pixel pathway}
\label{sec:partialhead}
JiT, %
as well as other recent work~\citep{pixeldit,dip,l2p} enable the pixel pathway throughout the entire generation process.
However, the additional pixel head can introduce substantial computational overhead.
This begs the question: \textbf{Is the pixel head necessary at all denoising steps?}
Since the pixel head is envisioned primarily as a component that generates the finer details in the image, we first look when the details emerge during the generative process and what predictions look like at different timesteps.

\begin{figure}
    \centering
    \includegraphics[width=0.8\linewidth]{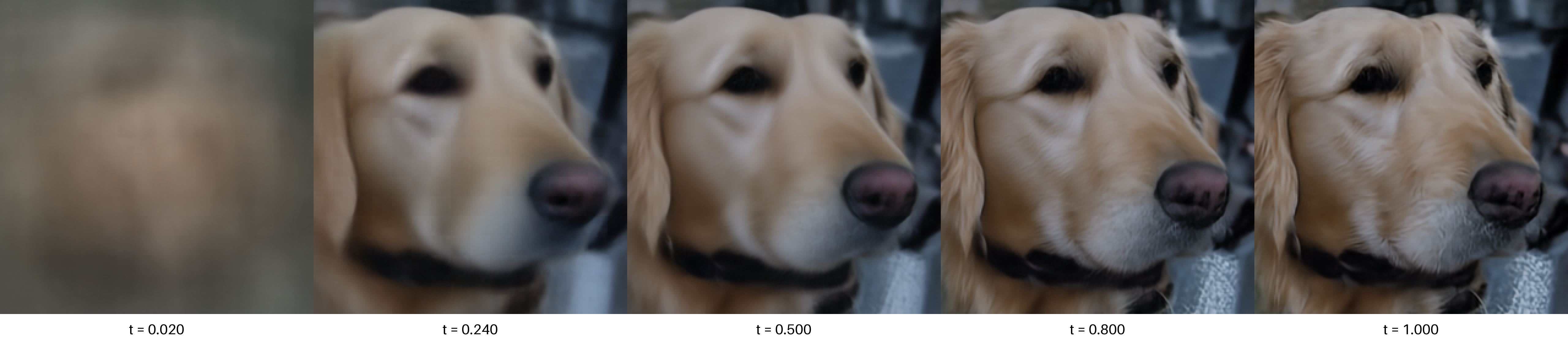}
    \caption{The predictions of the pretrained JiT-B/16 model at different timesteps $t$. Notice the smoothness of predictions at $t \leq 0.5$.}
    \label{fig:jit_xpredictions}
\end{figure}

At early stages of denoising, the low signal-to-noise ratio of $\x_t$ leaves substantial uncertainty about the underlying clean image $\x$.
The optimal solution for the model is then to predict the average over all clean images that can plausibly be reached from the given $\x_t$:
\begin{equation}
    \operatorname*{arg\,min}_{\xpred}
    \mathbb{E}\!\left[
        \|\xpred-\x\|_2^2
        \mid \x_t,t,c
    \right]
    =
    \mathbb{E}\!\left[\x\mid\x_t,t,c\right]  \enspace .
\end{equation}
Since multiple
clean images are compatible with
the noisy $\x_t$, this averaging suppresses details, yielding smooth and
blurry predictions, which is illustrated in \cref{fig:jit_xpredictions}.
As denoising progresses and the signal-to-noise ratio increases, the uncertainty about these details decreases, allowing increasingly fine structure to emerge in the predicted clean images.

\paragraph{Late-Stage Pixel Head.}
Inspired by the above observations, we develop a novel approach that trains two different pixel heads: (1) a lightweight patch-overlapping linear head $g_{\mathrm{linear}} (h, c)$, and (2) a convolutional head $g_{\mathrm{conv}} (h, \x_t, t, c)$ as described above.
During sampling, the linear head is active in earlier low-SNR denoising stages while the more expensive convolutional head is active in later high-SNR stages:

\begin{equation}
    \hat{\mathbf{x}} =
    \begin{cases}
        \xlin,
        & t \le t_a, \\[2pt]
        (1-s(t))\xlin + s(t)\xconv,
        & t_a < t < t_b, \\[2pt]
        \xconv,
        & t \ge t_b,
    \end{cases}
    \label{eq:head-schedule}
\end{equation}

where $
    \xlin = g_{\mathrm{lin}}(h, c)$ and $\xconv = g_{\mathrm{conv}}(h,\mathbf{x}_t,c,t)$
and $s(t)$ is a smooth interpolation function in the overlapping region.
Note that we train the convolutional head over the full timestep range and set the linear head’s loss weight to zero for \(t \ge 0.8\).

The \textbf{overlapping linear output layer} $g_{\text{linear}}$ used in the earlier stages is simply a linear layer, using the same implementation as JiT. However, instead of projecting backbone features $h$ to $\R^{16 \times 16 \times 3}$ that specifies values for each backbone patch, it also generates values for $K$ pixels surrounding the original patch. So, $g_{\text{linear}}: \R^{d_{\text{transformer}}} \rightarrow \R^{(16 + 2K) \times (16 + 2K) \times 3}$.
The final image is composed by smoothly interpolating between the different patches in the overlapping regions.
While the additional overhead of predicting overlapping regions with other patches is negligible, it helps smoothen predictions and prevent patch boundary discontinuities in early denoising steps.

\paragraph{Applicability of Late-Stage Pixel Pathway.}
Note that this dual-head approach is only feasible when training with the $x$-prediction objective.
When using $v$- or $\epsilon$-prediction, the model is expected to reproduce the exact original noise component present in the prediction target.
However, as \cite{jit} have shown, this is infeasible with a large-patch transformer with a linear output layer.

\section{Denoising correction}
\label{sec:samplingcorrection}
Before discussing the training method, we first take a detour to JiT's~\citep{jit} sampling procedure.
We observed that, in the original JiT codebase, the denominator $1-t$ used to recover the velocity from predicted $x$ is lower-bounded by $\epsclip=0.05$ during \textit{both} training and sampling.
While this stabilizes training by limiting the scaling of losses near $t=1$, clipping during sampling modifies the sampling trajectory.

Normally, the final Euler step completes the denoising process and predicts the final image $
    \x_1
    =
    \x_t+(1-t)\vpred
    =
    \hat{\x}_\theta$.

With the clipped denominator, however, the same update becomes
\begin{equation}
\begin{aligned}
    \x_1
    &= \x_t
    + \frac{1-t}{\max(1-t,\epsclip)}
      \left(\hat{\x}_\theta-\x_t\right)
    = \left(1-\frac{1-t}{\max(1-t,\epsclip)}\right)
       \x_t
    + \frac{1-t}{\max(1-t,\epsclip)}
      \hat{\x}_\theta \enspace.
\end{aligned}
\end{equation}
Whenever $1-t<\epsclip$, the sampler thus moves only a fraction $(1-t)/\epsclip$ toward the clean-image prediction, retaining part of the pre-terminal state.
This is equivalent to terminating at a nonzero noise level because the clipping preserves part of the noise still present in $\x_t$.
For example, with the default settings (50 steps, $\epsclip=0.05$), this gives $x_1 = 0.6 * x_{0.98} + 0.4 \xpred$.
Note that the previous denoising step (from $x_{0.96}$) is also affected by clipping.

\paragraph{Does clamping $\mathbf{(1-t)}$ in the velocity conversion during sampling improve FID and why?}
We compare the following three variants of the sampling procedure:
(1) the original clamped version ($\epsclip=0.05$),
(2) the unclamped version ($\epsclip=0.0$),
and (3) a version where we perform unclamped generation but subsequently add independent Gaussian noise to match the noise level of the original version.

\begin{figure}[t]
    \centering

    \begin{minipage}[t]{0.4\linewidth}
        \vspace{0pt}
        \centering
        \textbf{(a) Quantitative comparison}
        \vspace{3pt}

        \small
        \setlength{\tabcolsep}{3.5pt}
        \renewcommand{\arraystretch}{1.0}

    \setlength{\aboverulesep}{2pt}
    \setlength{\belowrulesep}{2pt}
        \begin{tabular}{@{}lcccc@{}}
            \toprule
            Model
            & $\epsclip$
            & $\sigma_m$
            & FID $\downarrow$
            & IS $\uparrow$ \\
            \midrule
            JiT-B/16
            & $0.05$ & $0$
            & $\mathbf{3.64}$ & $\mathbf{269.0}$ \\
            & $0$ & $0$
            & $5.43$ & $251.8$ \\
            & $0$ & $0.018$
            & $3.81$ & $263.1$ \\
            \midrule
            JiT-L/16
            & $0.05$ & $0$
            & $\mathbf{2.39}$ & $\mathbf{290.6}$ \\
            & $0$ & $0$
            & $4.45$ & $266.4$ \\
            & $0$ & $0.018$ & $2.72$ & $281.7$ \\
            \bottomrule
        \end{tabular}
    \end{minipage}
    \hfill
    \begin{minipage}[t]{0.59\linewidth}
        \vspace{0pt}
        \centering
        \textbf{(b--d) Qualitative comparison}
        \vspace{3pt}

        \begin{minipage}[t]{0.31\linewidth}
            \centering
            \includegraphics[width=\linewidth]
                {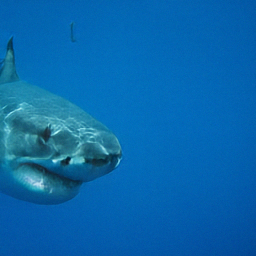}\\[-2pt]
            {\scriptsize\textbf{(b)} Clamp}
        \end{minipage}
        \hfill
        \begin{minipage}[t]{0.31\linewidth}
            \centering
            \includegraphics[width=\linewidth]
                {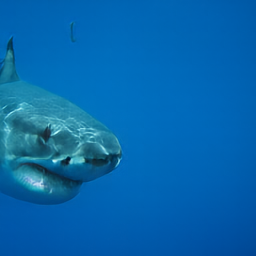}\\[-2pt]
            {\scriptsize\textbf{(c)} No clamp}
        \end{minipage}
        \hfill
        \begin{minipage}[t]{0.31\linewidth}
            \centering
            \includegraphics[width=\linewidth]
                {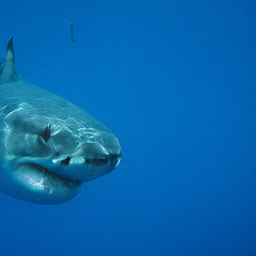}\\[-2pt]
            {\scriptsize\textbf{(d)} No clamp + Noise}
        \end{minipage}
    \end{minipage}

    \caption{
        Effect of denominator clamping during 50-step Heun sampling.
        \textbf{(a)} Quantitative results for pretrained JiT models with
        the original clamp ($\epsclip=0.05$), unclamped sampling, and
        matched-noise controls. CFG and EMA settings follow \citet{jit}.
        \textbf{(b--d)} JiT-L/16 samples with original clamp,
        no clamp, and no clamp followed by Gaussian noise with
        $\sigma_m=0.018$, respectively.
    }
    \label{fig:clamp_comparison}
\end{figure}

The results are shown for the pre-trained JiT-B/16 and L/16 models in \cref{fig:clamp_comparison}.
For JiT-B/16, removing the clamp increases FID from $3.63$ to $5.56$ (keeping the same CFG), confirming that the original sampling modification has a substantial effect on the measured image quality.
Interestingly, adding independent Gaussian noise to the fully denoised samples reduces FID to $3.84$, recovering approximately $89\%$ of the FID degradation caused by removing the clamp.
This supports the hypothesis that the original clamped sampler  benefits from preserving some random noise, rather than from producing a more reliable $x_1$ estimate.

\paragraph{Denoising correction.}
In our experiments, unless stated otherwise, we \emph{do not} clamp the denominator $(1-t)$ when computing $\vpred$ during sampling.
The sampler then follows the same trajectory as the equivalent $v$-prediction model and the final prediction has no residual noise.
Note that clamping is not needed for stability at sampling time.
An Euler step from $t$ to $t'$,
$
    \mathbf{x}_{t'}
    = \mathbf{x}_t + (t'-t)\,\hat{\mathbf{v}}_\theta
    = \mathbf{x}_t + \frac{t'-t}{1-t}\left(\hat{\mathbf{x}}_\theta - \mathbf{x}_t\right),
$
already scales $\vpred$ with $t' - t$, such that $\frac{t' -t }{1 - t} \leq 1$.

\section{From structure learning to endpoint refinement}
\label{sec:endpoint_training}

We use \textbf{two-stage training}.
This enables more efficient training by enabling the pixel pathway only in the second stage.
In the \textbf{first stage}, the transformer backbone is pre-trained using JiT's logit-normal timestep sampling ($ t \sim \operatorname{LogitNormal(-0.8,0.8^2)}$)\footnote{In practice, we use the pre-trained checkpoints provided by \cite{jit}.}.
Note that JiT's timestep sampler assigns extremely little probability mass do to the final part of the denoising trajectory ($\Pr(t>0.95) \approx 1.4\times10^{-6}$, about 35{,}000 times less likely than a uniform sampler), even though this region is where the model performs pixel-level refinement.
Therefore, in the \textbf{second stage}, where we train the pixel head and fine-tune the transformer backbone, uniform timestep sampling ($t\sim \mathcal{U}(0, 1)$ with a Soft-Min-SNR loss weighting is used:
\begin{equation}
\mathcal{L}_{\mathbf{v}}
=
\mathbb{E}_{\x\sim\mathcal{D},\,t\sim\mathcal{U}(0,1)}
\left[w_{\mathrm{soft}}(t)\left\|\xpred-\x\right\|_2^2\right],
\qquad
w_{\mathrm{soft}}(t)=\frac{\rho(t)}{\rho(t)+\gamma},
\qquad
\rho(t)=\frac{t^2}{(1-t)^2}.
\label{eq:soft-loss}
\end{equation}
Here, $\rho(t)$ denotes the signal-to-noise ratio at timestep $t$.
In essence, this preserves a weight close to $1$ near the clean endpoint ($t \rightarrow 1$) and downweighs the losses near the noisy endpoint ($t \rightarrow 0$). This is similar to the $v$-loss weighting but is bounded near $t=1$ and does not require clipping, while still preventing noisy gradients near $t \rightarrow 0$ from dominating the updates.
See \cref{sec:app:training} in the Appendix for more details.

\section{Related Work}
\label{sec:relatedwork}

Diffusion and score-based models generate images by reversing corruption~\citep{diffusion,ddpm,scoresde}, while flow matching and rectified flow learn noise-to-data vector fields~\citep{flowmatching,rectifiedflow}.
Latent diffusion~\citep{ldm} and DiT~\citep{dit} enable efficient synthesis in autoencoder representations.
\textbf{Pixel-space generation} avoids this compression: early examples include DDPM, ADM~\citep{adm}, and cascaded systems such as Imagen~\citep{imagen} and DeepFloyd IF~\citep{deepfloydif}.
Simple Diffusion~\citep{simplediffusion}, SiD2~\citep{sid2}, and HDiT~\citep{hdit} are based on efficient high-resolution architectures and training recipes; Edify~\citep{edify} introduces frequency-dependent Laplacian diffusion.

Several works have investigated \textbf{architectural choices for pixel diffusion models}.
JiT~\citep{jit} enables plain large-patch Transformers through clean-image prediction.
Its linear output projection reconstructs non-overlapping patches from individual backbone tokens, leaving boundary consistency and fine detail to patch-level representations.
Alternative detail mechanisms include PixNerd's patch-wise neural fields~\citep{pixnerd}, DiP's patch detailer~\citep{dip}, and PixelDiT's pixel pathway~\citep{pixeldit}.
The latter communicates across patches by compacting pixel features into patch tokens, applying global attention, and expanding back to pixels.
HyperDiT~\citep{hyperdit} connects fine-grained tokens to multilevel semantic features through cross-attention, while PixelU~\citep{pixelu} combines $x$-prediction with a U-shaped backbone and shallow-to-deep skips.
Our shallow convolutions instead directly mix neighboring spatial features across JiT's original patch boundaries.

Some recent work also investigate \textbf{improved training and post-training} for pixel diffusion models.
DeCo~\citep{deco} separates semantic and high-frequency modeling; PixelGen~\citep{pixelgen} adds perceptual supervision, and PixelREPA~\citep{pixelrepa} adapts representation alignment to JiT using transformed targets and a masked adapter.
Pixel-space supervision also improves post-training of latent models~\citep{pixelposttraining}, whereas L2P~\citep{l2p} and Z-Image-Pixel~\citep{l2pzimage} investigate conversion of latent models to pixel-space generation.
We instead fine-tune an already trained pixel-space JiT with an added refinement pathway, preserving the pretrained backbone as the starting point for learning local corrections.

Concerning \textbf{coarse-to-fine denoising},
PixelFlow~\citep{pixelflow}, Scale Space Diffusion~\citep{scalespace}, and Spectral Progressive Diffusion~\citep{spectralprogressive} reduce early computation through progressive resolution.
Spectral Forcing~\citep{spectralforcing} increases the input frequency bandwidth with denoising time, while Latent Forcing~\citep{latentforcing} coordinates latent and pixel noise schedules to establish structure before pixel detail.
PixelGen~\citep{pixelgen} restricts perceptual supervision to lower-noise timesteps.
These methods motivate timestep-dependent processing but use different mechanisms from our head switch: at fixed image resolution, an overlapping linear head handles early denoising, followed by a convolutional pixel head in later steps.
We study whether this late-only refinement retains the quality benefit of applying the added head throughout sampling.

\begin{table}[t]
    \centering
    \small
    \setlength{\abovecaptionskip}{5pt}
    \setlength{\belowcaptionskip}{0pt}
    \caption{Comparison of pixel-head variants for JiT-B and JiT-L.
    For Dual Head variants, GFLOPs are reported with the linear head and the convolutional head separately.
    Further details and breakdowns are provided in \cref{sec:app:timings}.
    }
    \label{tab:results:convjit}
    \renewcommand{\arraystretch}{1.12}
    \setlength{\tabcolsep}{7pt}

    \begin{tabular}{@{}l cccccc@{}}
        \toprule
        Method
        & Params.
        & GFLOPs
        & Time $\downarrow$
        & gFID $\downarrow$
        & sFID $\downarrow$
        & IS $\uparrow$ \\
        & (M) & (per FE) & (s/img) & & & \\
        \midrule

        \multicolumn{7}{@{}l}{\textbf{JiT-B}} \\
        \addlinespace[3pt]
        Pre-trained ($\epsilon = 0.05$)
            & 131.3 & 50.4 & 1.44 & 3.64 & 7.21 & 269.0\\
        Pre-trained ($\epsilon = 0.0$)
            & 131.3 & 50.4 & 1.44 & 4.61 & 8.68 & 282.4 \\
        Fine-tuned (100 epochs)
            & 131.3 & 50.4 & 1.44 & 4.29 & 7.21 & 275.7\\

        \addlinespace[3pt]
        Conv.\ pixel head
            & 132.9 & 60.9 & 1.74 & 3.42 & 5.49 & 264.3 \\
        Dual pixel head ($0.5 - 0.7$)
            & 135.4 & 50.8 / 60.9 & 1.55 & 3.41 & 5.71 & 269.3\\
        Dual pixel head ($t_{\text{thresh}} = 0.8$)
            & 135.4 & 50.8 / 60.9 & 1.49 & 3.42 & 6.38 & 270.7\\

        \midrule
        \multicolumn{7}{@{}l}{\textbf{JiT-L}} \\
        \addlinespace[3pt]
        Pre-trained ($\epsilon = 0.05$)
            & 459.1 & 175.9 & 3.08 & 2.39 & 6.28 & 290.6 \\
        Pre-trained ($\epsilon = 0.0$)
            & 459.1 & 175.9 & 3.08 & 3.51 & 9.15 & 324.6\\
        Fine-tuned (100 epochs)
            & 459.1 & 175.9 & 3.08 & 2.72 & 6.87 & 309.3 \\

        \addlinespace[3pt]
        Conv.\ pixel head
            & 460.6 & 186.8 & 3.27 & 2.27 & 5.35 & 308.2 \\
        Dual pixel head ($0.5 - 0.7$)
            & 464.5 & 176.4 / 186.8 & 3.21 & 2.24 & 5.44 & 309.8 \\
        Dual pixel head ($t_{\text{thresh}} = 0.8$)
            & 464.5 & 176.4 / 186.8 & 3.16 & 2.20 & 5.98 & 307.5 \\

        \bottomrule
    \end{tabular}
    \vspace{-5pt}
\end{table}

\section{Experiments}
\label{sec:experiments}

\subsection{Experimental Setup}
\label{sec:setup}

\paragraph{Training setup.}
For each experiment, we initialize the model from the pretrained JiT checkpoint and augment it with the randomly initialized layers introduced in \cref{sec:approach}.
In the first stage of fine-tuning (15 epochs unless otherwise specified), only the added pixel pathway is trained while the backbone remains frozen.
In the second stage of fine-tuning (85 epochs), the pretrained backbone is unfrozen and trained together with the pixel pathway.
The learning rate is $5 \times 10^{-5}$ for the pretrained backbone and $1 \times 10^{-4}$ for the newly introduced layers.
The global batch size is fixed at 256, while all remaining training hyperparameters follow the defaults of the original JiT implementation.
We do not use any additional losses, such as the representation alignment loss (REPA~\citep{repaloss}).

\paragraph{Evaluation setup.}
For trained models, we run a sweep on EMA and CFG settings. First, we sweep different combinations of the varied hyperparameters with 10k images generated for each setting. The best setting is selected and 50k images are generated, which are used to measure generator quality using FID and IS like previous work. Generally, we sweep CFG's in the range $2.0$ to $3.5$ and try it with both EMA weights (decays of $0.9996$ and $0.9999$).
During generation, we use JiT's default Heun~\citep{heun} sampler with 50 steps.

\subsection{Results}
\label{sec:results}
\cref{tab:results:convjit} shows the effect of introducing the pixel pathway and its late application. Qualitative results are shown in Appendix~\ref{sec:supp:qualitative_examples}.

\paragraph{Convolutional pixel pathway.}
Comparing the metrics for the pre-trained model with the ''Conv. pixel head'' rows for both JiT-B and JiT-L in \cref{tab:results:convjit} reveals that there is indeed improvement in image quality, as measured by FID.
In particular, sFID, which is more attuned to superficial image quality, shows a larger drop.
Interestingly, the post-trained original JiT model control is also further improved in post-training, but still doesn't match the original numbers.
In fact, the gFID pushes below the originally reported results from \cite{jit}.
The parameter count increase is negligible compared to total parameter count.
Generation runtime increases $21\%$ for JiT-B and $6.2\%$ for JiT-L.
Note that for both backbones, we used a small-sized pixel pathway with only 4 dense convolutions.

\paragraph{Late pixel pathway.}
The results for the dual pixel head rows reveal that there is little to no quality degradation when the convolutional pixel pathway is applied only in the last 20\% of the denoising trajectory.
Introducing the late pixel pathway with the second overlapping linear output head further increases the parameter count but results in a reduction in generation speed for both models.
Even though the difference is small for the L model since the small convolutional head is inexpensive compared to the total runtime of JiT-L, this difference can grow much larger if a more expensive pixel head is used.
For example, an architectural estimate for PixelDiT-XL gives 220 GFLOPs for their backbone and 90 GFLOPs for their pixel pathway.
This means that the pixel pathway consumes nearly 30\% of total compute.
With the late pixel pathway, this can be reduced to just around 8\% of total compute with the same head, which gives 23\% total savings in GFLOPs.

\subsection{Comparison}
\label{sec:comparison}
\begin{table}[t]
    \centering
    \small
    \setlength{\abovecaptionskip}{5pt}
    \setlength{\belowcaptionskip}{0pt}
    \caption{Comparison on ImageNet$@256$ for recent pixel-space generative models.
    For full comparison, see \cref{tab:fullcomparison} in \cref{sec:app:comparison}.
    For the H-model, we use AdaLN-based conditioning and wider pixel head (BW-Cond variant; see \cref{sec:app:headsize}) and $t \sim \operatorname{LogitNormal}(0, 1)$. The convolutional head is activated at $t=0.5$.
    }
    \label{tab:comparison}
\begin{tabular}{@{}llcccccc@{}}
    \toprule
    Method &  & REPA & Epochs & Params & NFE & gFID$\downarrow$ & IS$\uparrow$ \\
    \midrule
    Simple Diffusion & {\scriptsize \citep{simplediffusion}} & -- & 800 & 2.0B & -- & 2.44 & 256.3 \\
    PixelDiT-XL & \citep{pixeldit} & $\checkmark$ & 320 & 797M & $100 \times 2$ & 1.61 & 292.7 \\
    DeCo-XL/16  & \citep{deco}     & $\checkmark$ & 600 & 682M & $100 \times 2$ & 1.69 & 304.0 \\
    DiP-XL/16   & \citep{dip}      & $\checkmark$ & 600 & 631M & $100 \times 2$ & 1.79 & 281.9 \\
    PixelU-H/16 & \citep{pixelu}   & $\checkmark$ & 600 & 1.17B & $100 \times 2$ & 1.63 & 305.9 \\
    \midrule
    JiT-H/16    & \citep{jit}      & -- & 600 & 953M & $100 \times 2$ & 1.86 & 303.4 \\
    JiT-H/16 ($\epsilon = 0.0$)
        & Reproduction & -- & 600 & 953M & $100 \times 2$ & 2.34 & 335.4 \\
    \midrule
    JiT-H/16 + Dual Head
        & Ours & -- & 600+100 & 976M & $100 \times 2$ & 1.78 & 333.6 \\
    \bottomrule
\end{tabular}
\vspace{-5pt}
\end{table}

A comparison to previous work is shown in \cref{tab:comparison}. A full comparison is provided in \cref{tab:fullcomparison} in \cref{sec:app:comparison} in the Appendix.
JiT-H/16 obtains an FID of 2.34 with the corrected (unclamped) sampler.
Our method (JiT-H/16 + Dual Head, starting at $t=0.5$) obtains an FID of 1.78, beating the original JiT-H/16 and JiT-G/16 models.
Note that most other previous works used REPA losses to further improve their results.
Nevertheless, our numbers remain competitive and illustrate our main finding of improving a pretrained JiT checkpoint with a late pixel pathway extension.

\subsection{Ablations}
\label{sec:ablations}

\paragraph{Starting point of the late pixel pathway.}
We vary the inference-time transition while keeping the trained JiT-B checkpoint fixed (\cref{tab:results:transition}).
Restricting the pixel pathway to the final $20\%$ of the trajectory with a hard switch at $t=0.8$ yields a gFID of $3.42$, compared to $3.39$ for a smooth transition over $[0.5,0.6]$.
However, delaying the transition to $[0.9,0.99]$ increases gFID to $5.39$, suggesting that the pixel pathway requires sufficient time for late-stage refinement.

\begin{table}[t]
    \centering
    \small
    \setlength{\abovecaptionskip}{5pt}
    \setlength{\belowcaptionskip}{0pt}
    \caption{Transition ablation for JiT-B and JiT-L with Dual Pixel Heads on ImageNet
    $256\times256$, measured using gFID with 50k samples. Smooth transitions blend the linear and pixel
    heads over the indicated interval; hard switches activate the
    pixel head at the indicated timestep.
    }
    \label{tab:results:transition}
    \setlength{\tabcolsep}{4pt}
    \begin{tabular}{@{}lccccc c ccccc@{}}
    \toprule
    & \multicolumn{5}{c}{Gradual}
    &
    & \multicolumn{5}{c}{Hard} \\
    \cmidrule(lr){2-6} \cmidrule(lr){8-12}

    Method
    & $.50$--$.60$ & $.60$--$.70$ & $.70$--$.80$ & $.80$--$.90$ & $.90$--$.99$
    &
    & $.60$ & $.70$ & $.80$ & $.90$ & $.95$ \\
    \midrule

    JiT-B + Dual Pixel Head
    & 3.39 & 3.41 & 3.41 & 3.46 & 5.39
    &
    & 3.40 & 3.41 & 3.42 & 3.58 & 4.98 \\

    JiT-L + Dual Pixel Head
    & 2.24 & 2.23 & 2.21 & 2.20 & 3.55
    &
    & 2.24 & 2.22 & 2.20 & 2.27 & 3.24 \\
    \bottomrule
\end{tabular}
\end{table}

\begin{table}[t]
    \centering
    \small
    \setlength{\abovecaptionskip}{5pt}
    \setlength{\belowcaptionskip}{0pt}
    \caption{Ablation results on $\x_t$ skip connection (left) and residual pathway (right).
    }
    \label{tab:ablation:res}
    \setlength{\tabcolsep}{4pt}
    \renewcommand{\arraystretch}{0.95}

    \begin{tabular*}{0.485\linewidth}{
        @{\extracolsep{\fill}}lcc@{}
    }
        \toprule
        JiT-B + Dual Pixel Head & gFID $\downarrow$ & IS $\uparrow$ \\
        \midrule
        Baseline           & 3.41 & 269.3 \\
        No skip from $\x_t$ & 3.91 {\scriptsize (+0.50)} & 277.9 \\
        \bottomrule
    \end{tabular*}%
    \hfill
    \begin{tabular*}{0.485\linewidth}{
        @{\extracolsep{\fill}}lcc@{}
    }
        \toprule
        JiT-B + Conv Head (Residual) & gFID $\downarrow$ & IS $\uparrow$ \\
        \midrule
        Baseline        & 3.53 & 266.5 \\
        Backbone frozen & 4.20 {\scriptsize (+0.67)} & 283.6 \\
        \bottomrule
    \end{tabular*}
    \vspace{-5pt}
\end{table}

\paragraph{No skip from $\x_t$.}
In \cref{eq:pixelhead}, the pixel head receives the noisy pixel input $\x_t$ directly.
This shortcut essentially bypasses the original big-patch backbone while remaining conditioned on the large-patch backbone's features.
Related work (DiP, PixelDiT) discusses the benefit of explicit local modeling under a v-prediction approach but does not study whether direct access to $\x_t$ remains equally important under JiT's x-prediction parameterization.
Under JiT's manifold intuition, x-prediction avoids directly predicting the exact noise component present in the velocity target, potentially reducing the need for a high-frequency bypass.

The second row in \cref{tab:ablation:res} shows the effect of not feeding $\x_t$ directly to the pixel head. The FID degrades from 3.41 to 3.91. However, this is not as detrimental as disabling $(1-t)$ clipping (FID $4.61$), suggesting that some image quality can be recovered through architecture choices alone.
While we couldn't find similar ablations for $v$-prediction objectives in previous work, we expect that the FID drop in those cases will be more significant since $\x_t$ is important for predicting the exact noise pattern still present in $\x_t$.

\paragraph{Residual pixel pathway.}
We also experiment with a residual formulation for the pixel head, where it adds to the original prediction.
Third row of \cref{tab:ablation:res} shows that this also achieves good image quality and is thus a viable alternative to simply replacing the original output layer.
The fourth row reports results for using the residual formulation but keeping the backbone frozen.
The relatively poor result (FID $4.20$) indicates that original predictions can not be easily corrected with a simple post-hoc addition and requires backbone fine-tuning.

\paragraph{Additional ablations.}
We present additional ablations and analysis in the Appendix.
\cref{sec:app:headsize} presents ablations on the pixel head design, including its size.
\cref{sec:app:training} presents an analysis on the effect of timestep sampling distributions and loss weights in the fine-tuning stage.

\section{Limitations}
\label{sec:limitations}
Our study prioritizes controlled comparisons with the pretrained JiT baselines in order to improve the understanding of pixel pathways. Our best results can likely be further improved by using more advanced training techniques and higher-capacity pixel heads.
Some open questions remain.
We do not establish the optimal point during training at which to introduce the pixel pathway during training.
Our exploration of alternative pixel head designs is limited, leaving the potential benefits of higher-capacity designs unresolved.
Scaling beyond the evaluated model sizes and to text-to-image models and adapting pretrained latent-space diffusion models are other directions for future work.

\section{Conclusion}
\label{sec:conclusion}
In this work, we showed that fine-tuning a simple convolutional pixel head on top of pretrained JiT models can significantly improve image quality.
In addition, we find that the pixel pathway is not necessary throughout the entire denoising process, and can be activated as late as in the last 20\% of the denoising trajectory, which reduces computation requirements and further illustrates the usefulness of the $x$-prediction objective for pixel-space modeling.
We hope our findings inspire future work in this area and lead to further improvements in pixel-space diffusion transformers.

\clearpage

\subsection*{AI use statement}

AI-based tools were used during the preparation of this work to assist with writing, in particular the formatting and improvement of result tables, diagrams, drafting/rephrasing some paragraphs and reviewing written text. All generated tables, diagrams and text were manually reviewed and adjusted or reformulated to ensure correctness and improve clarity of presentation.
The core research ideas and approach were developed independently, with AI-assisted search being used to verify and find certain details in literature. Experimental settings were determined based on previous work.
All experiments have been designed, executed and interpreted manually by the authors.
The authors assume full responsibility for the final manuscript.

AI-based tools were also used to write several ablation and analysis scripts (e.g. convolutional head transition sweep and progression of $x_0$ predictions) and modify previous manually written code for additional features.
Core parts of the code (training, model, inference) are based on code previously provided by \cite{jit}, and were modified mostly manually, except for certain secondary features (e.g. reproducible seeds per image).
AI assisted during manual code review to double-check correctness and identify possible implementation errors.
All AI-generated code was manually reviewed by the main author.

\subsection*{Ethics statement}

This work does not involve human participants, private or sensitive information, or experiments that present foreseeable risks to individuals. Our evaluation is based entirely on publicly available datasets and pretrained models, used under their applicable licenses and usage conditions. We follow established research practices in the design, evaluation, and reporting of our experiments, with an emphasis on accurate and transparent presentation of the results. We have not identified any ethical concerns specific to the proposed approach beyond the broader considerations inherent to the datasets and models on which it builds.

\subsection*{Reproducibility statement}

Our experiments use ImageNet and publicly available pretrained JiT checkpoints. We describe the architectural modifications in Section 2, the sampling correction in Section~\ref{sec:samplingcorrection}, and the training objective in Section~\ref{sec:endpoint_training}. Section~\ref{sec:setup} specifies the fine-tuning schedule, learning rates, batch size, and evaluation protocol, including EMA and classifier-free guidance sweeps. Computational cost breakdowns are provided in Appendix~\ref{sec:app:timings}. We will release our code and model checkpoints.\\
Anonymized code is provided at {\url{https://anonymous.4open.science/r/JiT-BD8B}}.

\bibliography{iclr2027_conference}
\bibliographystyle{iclr2027_conference}

\clearpage
\appendix

\section{Full Comparison}
\label{sec:app:comparison}

\begin{table}[ht]
    \centering
    \small
    \caption{Full comparison on $256\times 256$ generation on ImageNet.
    NFE values with $\times 2$ denote nominal counts with classifier-free
    guidance; guidance intervals may reduce the actual count.
    RAE uses AutoGuidance with an additional guiding model.
    }
    \begin{tabular}{@{}llccccc@{}}
        \toprule
        Method & & Epochs & Params & NFE & gFID$\downarrow$ & IS$\uparrow$ \\
        \midrule
        \multicolumn{7}{@{}l}{\textit{Latent Generative Models}} \\
        \midrule
        DiT-XL
            & {\small \citep{dit}}
            & 1400 & 675M + 49M & $250 \times 2$ & 2.27 & 278.2 \\
        SiT-XL
            & \citep{sit}
            & 1400 & 675M + 49M & $250 \times 2$ & 2.06 & 277.5 \\
        MaskDiT
            & \citep{maskdit}
            & 1600 & 675M + 49M & $79 \times 2$ & 2.28 & 276.6 \\
        REPA
            & \citep{repaloss}
            & 800 & 675M + 49M & $250 \times 2$ & 1.42 & 305.7 \\
        LightningDiT
            & \citep{lightningdit}
            & 800 & 675M + 41M & $250 \times 2$ & 1.35 & 295.3 \\
        RAE-XL
            & \citep{rae}
            & 800 & 839M + 415M & -- & \textbf{1.13} & 262.6 \\
        \midrule
        \multicolumn{7}{@{}l}{\textit{Pixel Generative Models}} \\
        \midrule
        StyleGAN-XL
            & \citep{stylegan}
            & -- & -- & -- & 2.30 & 265.1 \\
        Simple Diffusion
            & {\scriptsize \citep{simplediffusion}}
            & 800 & 2.0B & -- & 2.44 & 256.3 \\
        PixelFlow-XL
            & \citep{pixelflow}
            & 320 & 677M & Adaptive & 1.98 & 282.1 \\
        PixNerd-XL
            & \citep{pixnerd}
            & 320 & 700M & $100 \times 2$ & 1.93 & 298.0 \\
        PixelDiT-XL
            & \citep{pixeldit}
            & 320 & 797M & $100 \times 2$ & 1.61 & 292.7 \\
        PixelGen-XL/16
            & \citep{pixelgen}
            & 160 & 676M & $100 \times 2$ & 1.83 & 293.6 \\
        DeCo-XL/16
            & \citep{deco}
            & 600 & 682M & $100 \times 2$ & 1.69 & 304.0 \\
        DiP-XL/16
            & \citep{dip}
            & 600 & 631M & $100 \times 2$ & 1.79 & 281.9 \\
        PixelREPA-H/16
            & \citep{pixelrepa}
            & 600 & 953M & $100 \times 2$ & 1.81 & 317.2 \\
        AsymFlow-H/16
            & \citep{asymflow}
            & 600 & 953M & $100 \times 2$ & 1.57 & -- \\
        {\scriptsize FrequencyBooster-H/16}
            & \citep{frequencybooster}
            & 320 & 1.1B & $100 \times 2$ & 1.60 & 321.0 \\
        PixelU-H/16
            & \citep{pixelu}
            & 600 & 1.17B & $100 \times 2$ & 1.63 & 305.9 \\
        PixSGR-H/16
            & \citep{pixsgr}
            & 320 & 875M & $100 \times 2$ & \textbf{1.51} & 311.0 \\
        \midrule

        \multicolumn{7}{@{}l}{\textit{JiT Models}} \\
        \midrule
        JiT-H/16
    & \citep{jit}
    & 600 & 953M & $100 \times 2$ & 1.86 & 303.4 \\
        JiT-G/16
            & \citep{jit}
            & 600 & 2B & $100 \times 2$ & 1.82 & 292.6 \\
JiT-H/16 ($\epsilon = 0.0$)
    & Reproduction
    & 600 & 953M & $100 \times 2$ & 2.34 & 335.4 \\
JiT-H/16 ($\epsilon = 0.05$) & Reproduction & 600 & 953M & $100 \times 2$ & 1.88 & 302.7 \\
\midrule
JiT-H/16 + Dual Head
    & Ours
    & 600+100 & 976M & $100 \times 2$ & \textbf{1.78} & 333.6 \\
        \bottomrule
    \end{tabular}
    \label{tab:fullcomparison}
\end{table}

A full comparison is provided in \cref{tab:fullcomparison}.

\section{Timestep sampling and loss weighting}
\label{sec:app:training}

\paragraph{Sparse endpoint coverage.}
JiT samples training timesteps $t$ from a logit-normal distribution,
    $t \sim \operatorname{LogitNormal(-0.8,0.8^2)}$.
This distribution assigns extremely little probability to the final part of the denoising path:
\begin{equation}
    \Pr(t>0.9)
    \approx
    9.0\times10^{-5},
    \qquad
    \Pr(t>0.95)
    \approx
    1.4\times10^{-6}.
\end{equation}
This means that JiT's logit-normal sampler visits $t>0.95$ approximately $35{,}000$ times less frequently than a uniform sampler.
Although these intervals constitute only a small portion of the flow, they correspond to low-noise states close to the data endpoint, where the model must perform accurate late-stage refinement.
Undersampling in this region can be especially consequential in pixel-space generative models where residual prediction errors directly affect fine image details.

Other works~\citep{pixeldit,dip} are better in this regard since they use logit-normal timestep sampling with $\mathcal{N}(0, 1)$, which doesn't suppress the near-clean region ($t > 0.9)$ as strongly.

Stronger endpoint coverage can be achieved by using using uniformly sampled timesteps ($t \sim \mathcal{U}(0, 1)$).

\begin{figure}[t]
    \centering
    \includegraphics[width=0.75\linewidth]{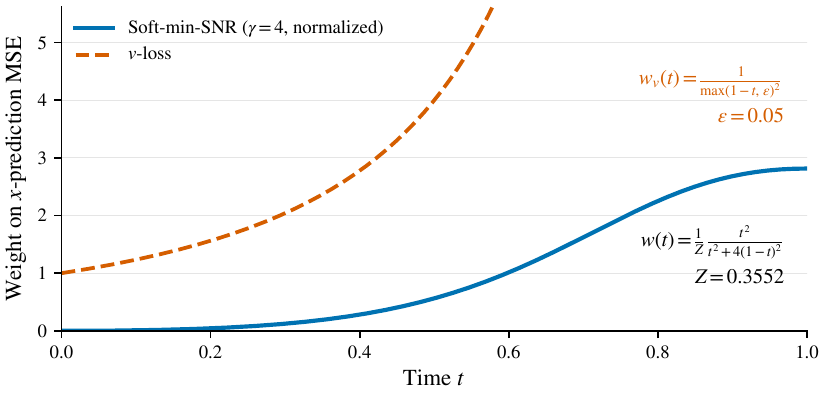}
    \caption{Weights applied to the $x$-prediction MSE: normalized soft-min-SNR ($\gamma=4$) and $v$-loss. The $v$-loss curve is clipped at the plot boundary.}
    \label{fig:lossweightgraphs}
\end{figure}

\begin{figure}[t]
    \centering
    \includegraphics[width=0.75\linewidth]{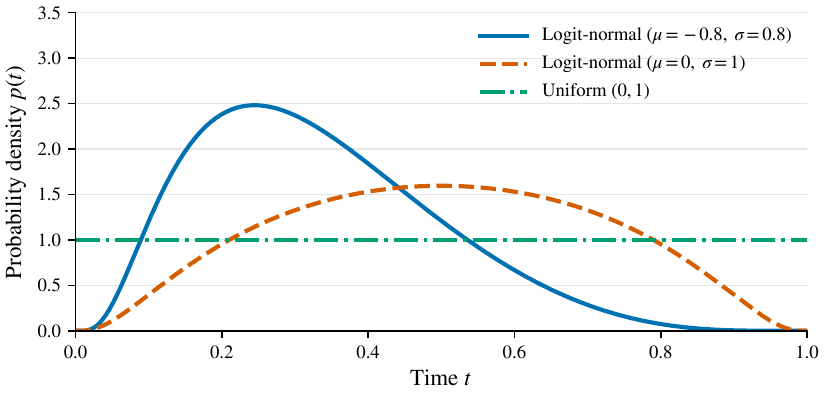}
    \caption{Timestep sampling densities for logit-normal distributions with $(\mu,\sigma)=(-0.8,0.8)$ and $(0,1)$, and uniform sampling on $[0,1]$.}
    \label{fig:timestepdistributions}
\end{figure}

To understand the effect of timestep sampling distribution and loss weighting, in this work, we experiment with these three options:
\begin{enumerate}
    \item original $ t \sim \operatorname{LogitNormal(-0.8,0.8^2)}$ with v-loss (\cref{eq:vloss}).
    \item standard $ t \sim \operatorname{LogitNormal(0,1)}$ with v-loss.
    \item $t \sim \mathcal{U}(0, 1)$ with Soft-Min-SNR weighting
\end{enumerate}
The timestep sampling distributions and the loss weights are plotted in \cref{fig:timestepdistributions} and \cref{fig:lossweightgraphs}, respectively.

\paragraph{Loss weighting.}

A model that predicts the clean endpoint $x$ can be trained using a weighted MSE loss:
\begin{align}
    \mathcal{L}
    =
    w(t)
    \left\|
        \xpred-\x
    \right\|_2^2.
    \label{eq:loss}
\end{align}

When converting the $v$-loss MSE objective $\mathcal{L} = \mathbb{E}[\left\|\hat{v}_{\theta}-\mathbf{v}\right\|_2^2]$ to $x$-prediction, the following loss emerges:

\begin{equation}
\mathcal{L}_{\mathbf{v}}
=
\underset{\x\sim\mathcal{D}, t\sim\mathcal{T}}{\mathbb{E}}
\left[
w(t)\left\|\xpred-\x\right\|_2^2
\right],
    \qquad
    w_{\text{v-loss}}(t)=\frac{1}{(1-t)^2}.
    \label{eq:vloss}
\end{equation}

In practice, the denominator should be clipped to ensure numerical stability.

\paragraph{Soft-Min-SNR endpoint refinement.}
Uniform sampling alone does not ensure that endpoint errors receive sufficient optimization pressure.
Direct clean-image prediction is more difficult at low $t$, where there is considerable noise in the image. This leaves the model with high uncertainty about prediction target and therefore the regression errors become large.
On the other hand, because of the low noise, errors near $t=1$ are typically much smaller since the image has already been largely formed.
With uniformly weighted clean-image objective, the errors near $t=0$ may therefore dominate the updates, slowing convergence near $t=1$.
To remedy this, timestep-dependent error weighting can be applied that suppresses noisy regions near $t=0$.

We combine uniform sampling with a Soft-Min-SNR-weighted loss weighting.
For JiT's linear interpolation path, we define
\begin{equation}
    \operatorname{SNR}(t)
    =
    \frac{t^2}{(1-t)^2}
\end{equation}
and optimize
\begin{equation}
    \mathcal{L}_{\mathrm{soft}}
    =
    \mathbb{E}_{t\sim\mathcal{U}(0,1)}
    \left[
        \frac{w(t)}{Z}
        \left\|
            \hat{\x}_\theta(\x_t,t)
            -
            \x
        \right\|_2^2
    \right],
\end{equation}
where
\begin{equation}
    w(t)
    =
    \frac{\operatorname{SNR}(t)}
         {\operatorname{SNR}(t)+\gamma},
    \qquad
    Z
    =
    \mathbb{E}_{t\sim\mathcal{U}(0,1)}
    \left[w(t)\right].
\end{equation}
The normalization $Z$ keeps the expected loss weight constant, preventing the change of objective from implicitly modifying the effective learning rate.

This weighting suppresses the large and noisy errors near $t=0$ while increasing the relative contribution of errors near $t=1$.
Compared to the inverse-square weighting induced by JiT's $\mathbf{v}$-loss, it remains bounded as $t\rightarrow1$ and doesn't require denominator clipping.
The same functional form has appeared as Soft-Min-SNR weighting in pixel-space diffusion~\citep{hdit}.

\subsection{Ablation}

\begin{table}[ht]
    \centering
    \caption{Comparison of timestep sampling and loss weighting on ImageNet $256\times256$ using the dual-head JiT-B/16 model with the B head. We select the lowest FID from final-epoch CFG/EMA sweeps from JiT's implementation, evaluated on 50,000 images and report IS from the same evaluation. All selected results use 50-step Heun sampling and decoder-transition start 0.5 and end 0.7.}
    \label{tab:timestep_loss_ablation}
    \small
    \setlength{\tabcolsep}{6pt}
    \begin{tabular}{llrr}
        \toprule
        Timestep distribution & Loss weighting & FID $\downarrow$ & IS $\uparrow$ \\
        \midrule
        Logit-normal ($\mu=-0.8$, $\sigma=0.8$) & $v$-loss & 3.65 & 282.86 \\
        Logit-normal ($\mu=0$, $\sigma=1$) & $v$-loss & 3.41 & 274.04 \\
        Uniform$^{*}$ & $v$-loss & 4.01 & 285.48 \\
        Uniform & Soft-min-SNR ($\gamma=4$) & 3.41 & 269.34 \\
        \bottomrule
    \end{tabular}
\end{table}

\cref{tab:timestep_loss_ablation} shows a comparison of timestep sampling and loss weighting using the dual-head JiT-B/16 model with the B head.

\section{Additional Pixel Head Ablations}
\label{sec:app:headsize}
\cref{tab:head_configuration_ablation} shows several ablations on the pixel head.

\begin{table}[ht!]
    \centering
    \caption{Head configuration ablation on ImageNet $256\times256$.
    We report FID and Inception Score (IS) on 50,000 generated images
    using 50-step Heun sampling. For each configuration, we select the
    lowest-FID evaluation from CFG/EMA sweeps using JiT's evaluator, and report its corresponding IS.
    Width and depth specify the head channel width and number of blocks.
    Head parameters (millions) include all weights and biases in the
    convolutional or ResNet head, excluding the backbone and overlap head.
    Head GFLOPs count convolution and matrix-multiplication operations per
    image for one forward pass.
    Head share is relative to the full forward pass.
    All runs use uniform timestep sampling and soft-min-SNR loss
    ($\gamma=4$).}
    \label{tab:head_configuration_ablation}
    \small
    \setlength{\tabcolsep}{3pt}
    \begin{tabular}{llrrrrrrr}
        \toprule
        Backbone & Head & Width & Depth & Head params (M) & Head GFLOPs & Share (\%) & FID $\downarrow$ & IS $\uparrow$ \\
        \midrule
        B/16 & T & 64 & 2 & 3.36 & 6.22 & 10.9 & 3.71 & 274.15 \\
        B/16 & B & 64 & 4 & 3.53 & 10.79 & 17.5 & 3.41 & 269.34 \\
        B/16 & B-Cond & 64 & 4 & 6.77 & 15.62 & 23.5 & 3.46 & 273.43 \\
        B/16 & BW & 128 & 4 & 7.71 & 39.83 & 43.9 & 3.28 & 272.28 \\
        B/16 & BW-Cond & 128 & 4 & 14.30 & 55.94 & 52.4 & 3.37 & 281.83 \\
        B/16 & L & 128 & 6 & 8.37 & 58.08 & 53.3 & 3.51 & 282.14 \\
        B/16 & ResNet-M & 96 & 2 & 5.22 & 13.36 & 20.8 & 3.69 & 270.17 \\
        \midrule
        L/16 & B & 64 & 4 & 4.59 & 11.32 & 6.0 & 2.24 & 309.83 \\
        L/16 & B-Cond & 64 & 4 & 8.90 & 16.69 & 8.6 & 2.18 & 308.18 \\
        L/16 & ResNet-M & 96 & 2 & 6.82 & 14.17 & 7.4 & 2.30 & 306.00 \\
        \bottomrule
    \end{tabular}
\end{table}

\paragraph{Head Size.}
For head sizes for the main ConvNeXt variant, we experimented with T(iny), B(ase), B(ase)-W(ide) and L(arge).
Each block contains only one dense $3\times 3$ convolution, followed by two $1\times 1$ convolutions.
Depth in the table refers to the number of such blocks. From the results, we see that the B and BW variants provide best performance, with T being too small.
    ResNet-M uses two residual blocks, each with two $3\times3$ convolutions.

\paragraph{Conditioning.}
We also experimented with using backbone features to condition the convolutional blocks using AdaLN modulation.
These are reported as "Cond" variants.
From our exploration, they do not appear to improve performance.

\paragraph{ResNet vs ConvNeXt.}
Finally, we also experimented with using ResNet blocks, reported as ResNet. We matched the receptive fields by using two blocks, each containing two convolutions.
From the results it appears that they do not perform better than our main ConvNeXt-like variant.

\section{Computational Cost Breakdown of ConvDual Models}
\label{sec:app:timings}

We first describe the computational resources used in our experiments. Then we report the computational cost of ConvDual-B/16,
ConvDual-L/16, and ConvDual-H/16 at an image resolution of $256 \times 256$.

\subsection{Compute Resources}
\label{sec:app:compute_resources}
We conducted large-scale training on single nodes located in a computing cluster equipped with four NVIDIA H100 GPUs and two AMD EPYC 9454 (48 core) CPUs per node. Initial experimentation and debugging were performed on a server with two 32-core AMD EPYC 7452 CPUs and four NVIDIA A40 GPUs (48 GB of VRAM each), running Ubuntu 20.04.6 LTS.

\subsection{Cost per network evaluation}

\begin{table}[ht]
    \centering
    \small
    \setlength{\tabcolsep}{5pt}
    \caption{
        Computational cost per network forward pass.
        Values are GFLOPs; percentages indicate the fraction
        of the corresponding row's total.
        ``Heads'' includes all active output heads.
        One CFG-guided evaluation requires two network forwards
        and therefore doubles all listed costs.
    }
    \label{tab:convdual_forward_flops}
    \begin{tabular}{llrrr}
        \toprule
        Model & Active heads & Total & Backbone & Heads \\
        \midrule
        B/16 & Linear
             & 50.82 & 50.14 (98.66\%) & 0.68 (1.34\%) \\
             & Conv
             & 60.93 & 50.14 (82.29\%) & 10.79 (17.71\%) \\
             & Both
             & 61.61 & 50.14 (81.38\%) & 11.47 (18.62\%) \\
        \midrule
        L/16 & Linear
             & 176.39 & 175.48 (99.48\%) & 0.91 (0.52\%) \\
             & Conv
             & 186.81 & 175.48 (93.94\%) & 11.32 (6.06\%) \\
             & Both
             & 187.72 & 175.48 (93.48\%) & 12.24 (6.52\%) \\
        \midrule
        H/16 & Linear
             & 364.52 & 363.38 (99.69\%) & 1.14 (0.31\%) \\
             & Conv
             & 423.61 & 363.38 (85.78\%) & 60.23 (14.22\%) \\
             & Both
             & 424.75 & 363.38 (85.55\%) & 61.37 (14.45\%) \\
        \bottomrule
    \end{tabular}
\end{table}
Table~\ref{tab:convdual_forward_flops} reports the cost of one
network forward pass, before classifier-free guidance (CFG).
When both output heads are active, they share a single backbone
evaluation.

The linear head costs only $0.682$ GFLOPs for ConvDual-B and
$0.910$ GFLOPs for ConvDual-L, compared with $10.788$ and
$11.325$ GFLOPs, respectively, for the convolutional head.
Consequently, evaluating both heads is only slightly more
expensive than evaluating the convolutional head alone.

\subsection{Cost per generated image.}

\begin{table}[ht]
    \centering
    \small
    \setlength{\tabcolsep}{4pt}
    \caption{
        Estimated computation per generated image using
        50 sampling steps with Heun, a final Euler step, and CFG.
        The convolutional head variant is shown below each model.
        Values are TFLOPs; percentages give each component's
        share of the total.
        Head costs aggregate all active head evaluations
        over the sampling trajectory.
    }
    \label{tab:convdual_image_flops}
    \begin{tabular}{llrrr}
        \toprule
        Model & Head schedule & Total & Backbone & Heads \\
        \midrule
        B/16 & Conv throughout
             & 12.1 & 9.9 (82.3\%) & 2.1 (17.7\%) \\
        {\scriptsize Head: B}
             & Blend $0.5$--$0.7$
             & 11.1 & 9.9 (89.8\%) & 1.1 (10.2\%) \\
             & Switch at $0.8$
             & 10.5 & 9.9 (94.8\%) & 0.5 (5.2\%) \\
        \midrule
        L/16 & Conv throughout
             & 37.0 & 34.7 (93.9\%) & 2.2 (6.1\%) \\
        {\scriptsize Head: B}
             & Blend $0.5$--$0.7$
             & 36.0 & 34.7 (96.6\%) & 1.2 (3.4\%) \\
             & Switch at $0.8$
             & 35.3 & 34.7 (98.3\%) & 0.6 (1.7\%) \\
        \midrule
        H/16 & Conv throughout
             & 83.9 & 71.9 (85.8\%) & 11.9 (14.2\%) \\
        {\scriptsize Head: BW-Cond}
             & Blend $0.5$--$0.7$
             & 77.9 & 71.9 (92.4\%) & 5.9 (7.6\%) \\
             & Switch at $0.8$
             & 74.4 & 71.9 (96.7\%) & 2.5 (3.3\%) \\
        \bottomrule
    \end{tabular}
\end{table}

We use the same sampling configuration as the other experiments: 50 denoising steps, with Heun updates for the
first 49 steps and an Euler update for the final step.
This requires 99 guided function evaluations.
Each guided evaluation performs a conditional and an
unconditional network forward, yielding 198 network
forwards per generated image.

We compare three head schedules as sampling time increases
from $t=0$ to $t=1$:
(i) the convolutional head throughout;
(ii) a transition from the linear overlapping head to the
convolutional head over $t \in [0.5,0.7]$; and
(iii) a hard switch from linear to convolutional predictions
at $t=0.8$.
For the transition schedule, only the linear head is used
at $t \leq 0.5$, both heads are used at
$0.5 < t < 0.7$, and only the convolutional head is used
at $t \geq 0.7$.
For the hard switch, the convolutional head is used at
$t \geq 0.8$.

Relative to using the convolutional head throughout,
the transition schedule reduces total computation by
$8.34\%$, $2.78\%$, and $7.13\%$ for ConvDual-B,
ConvDual-L, and ConvDual-H, respectively.
The hard switch at $t=0.8$ increases these savings to
$13.24\%$, $4.45\%$, and $11.27\%$, respectively.
The backbone cost remains constant across schedules;
all computational savings come from changing which
output heads are evaluated.

\clearpage

\subsection{Qualitative examples}
\label{sec:supp:qualitative_examples}

\begin{figure}[h]
    \centering
    \includegraphics[width=0.97\linewidth]{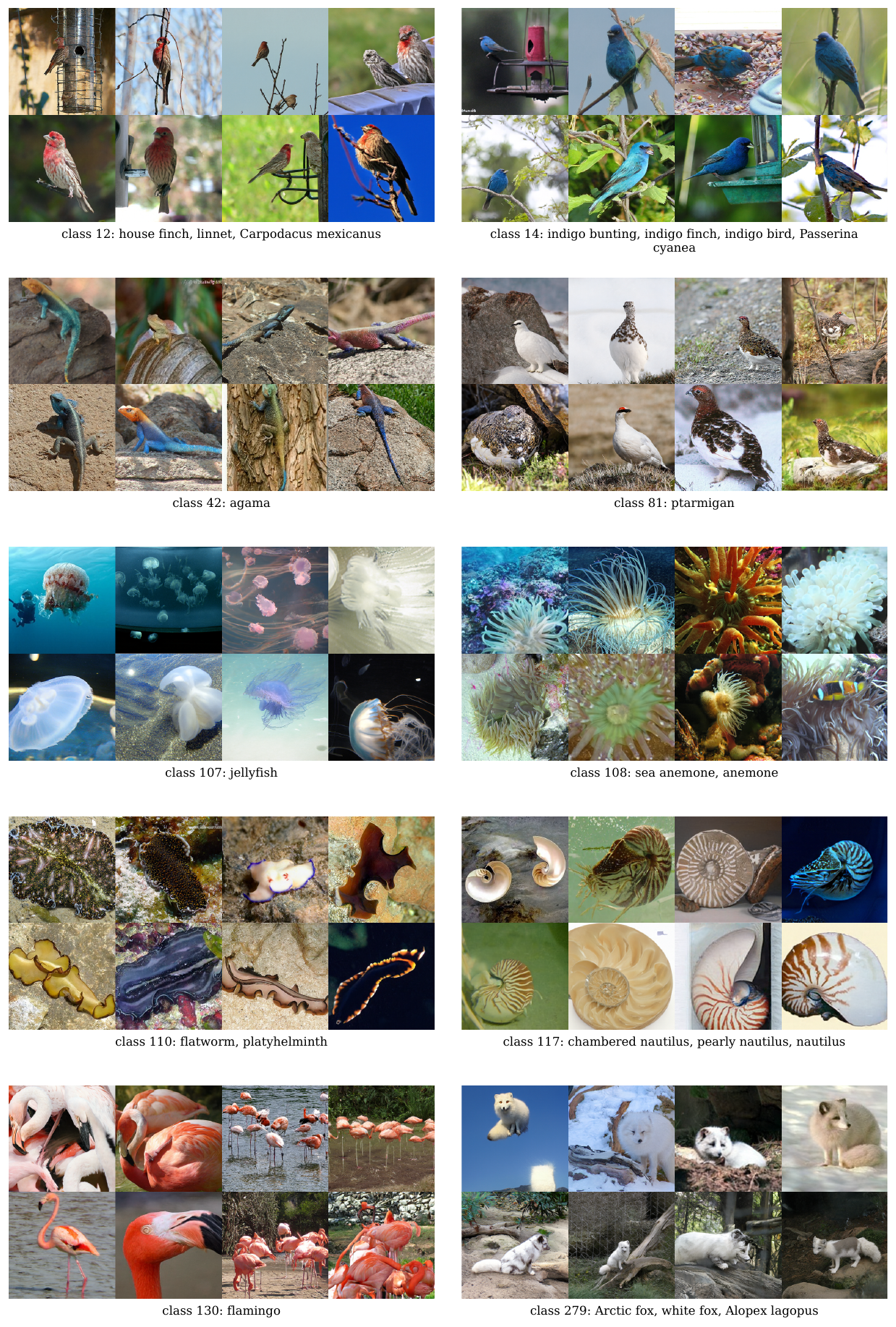}
  \caption{Qualitative samples on ImageNet $256\times256$, conditioned on the indicated classes, generated with JiT-H/16 + Dual Head and CFG of 2.8.}
  \label{fig:supp:qualitative-1}
\end{figure}

\begin{figure}[h]
    \centering
    \includegraphics[width=0.97\linewidth]{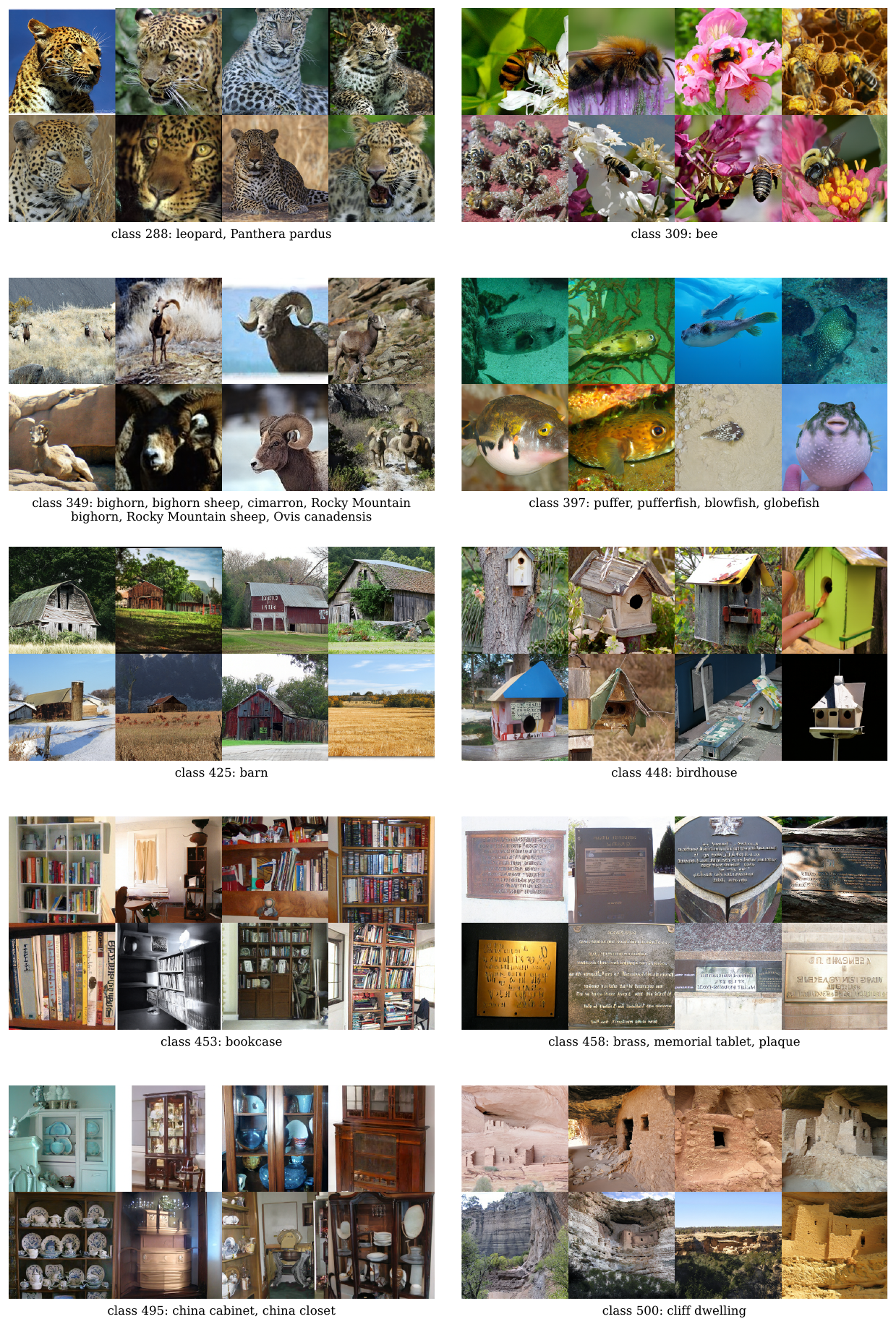}
    \caption{Qualitative samples on ImageNet $256\times256$, conditioned on the indicated classes, generated with JiT-H/16 + Dual Head and CFG of 2.8.}
  \label{fig:supp:qualitative-2}
\end{figure}

\begin{figure}[h]
    \centering
    \includegraphics[width=0.97\linewidth]{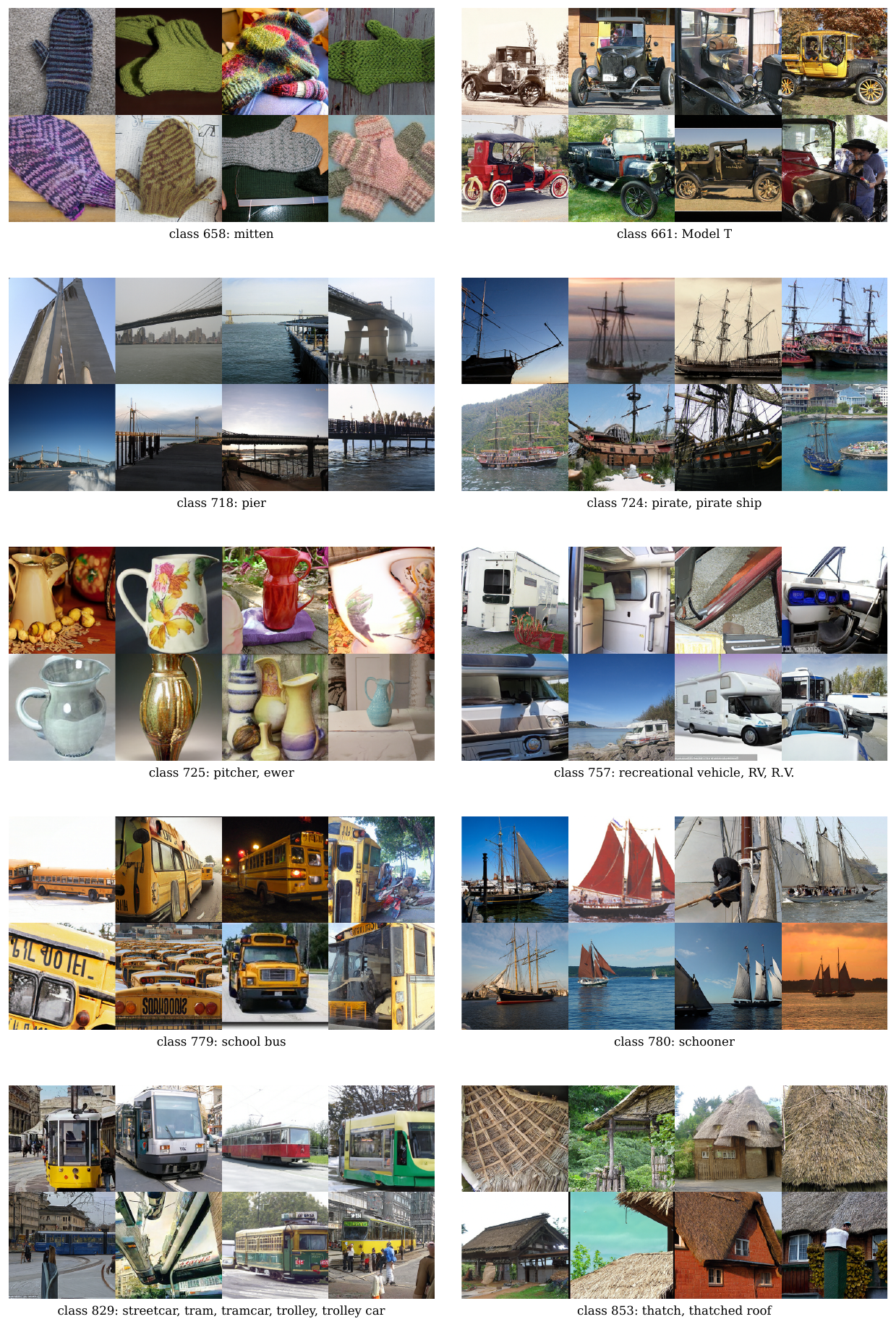}
  \caption{Qualitative samples on ImageNet $256\times256$, conditioned on the indicated classes, generated with JiT-H/16 + Dual Head and CFG of 2.8.}
  \label{fig:supp:qualitative-3}
\end{figure}

\begin{figure}[h]
    \centering
    \includegraphics[width=0.97\linewidth]{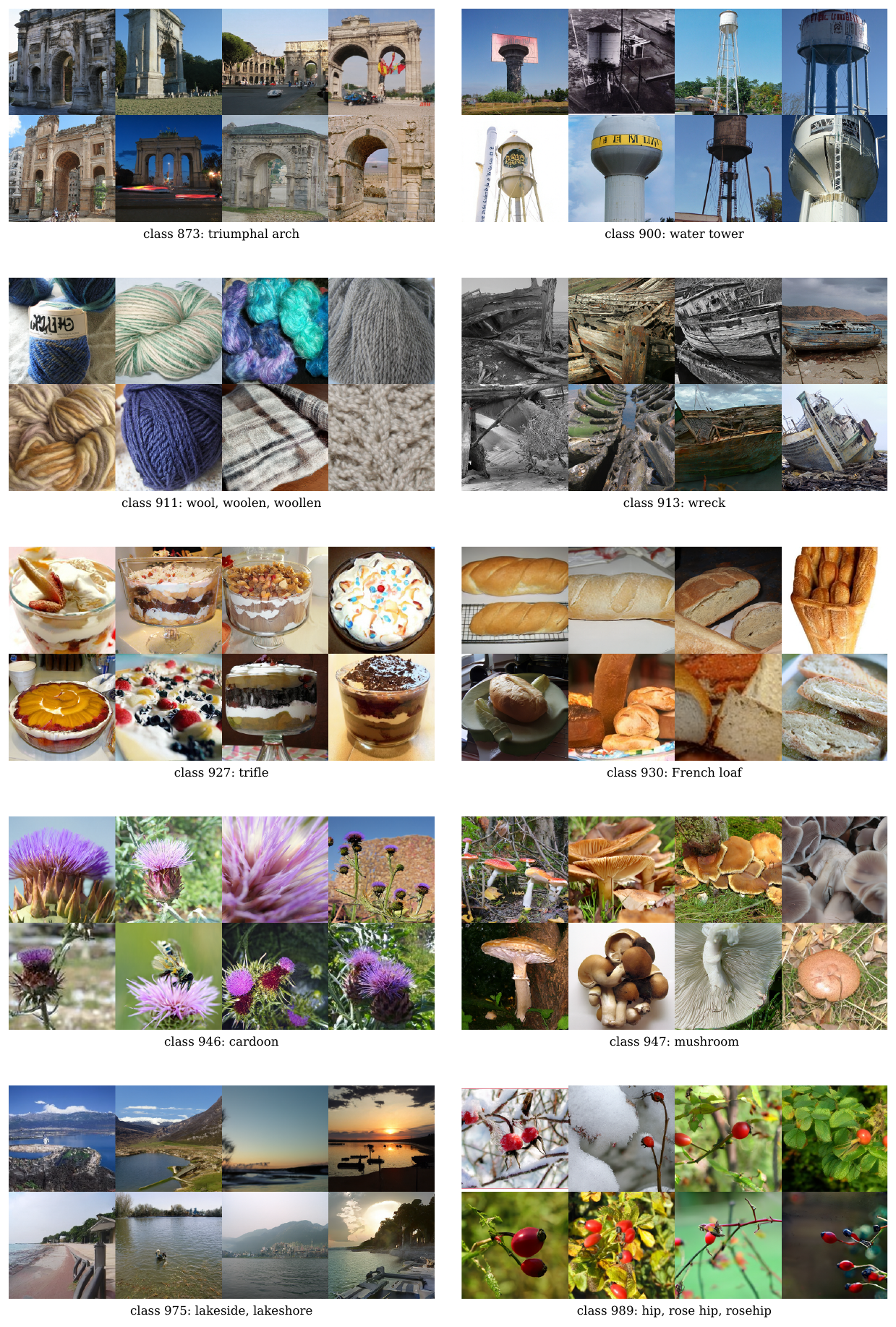}
  \caption{Qualitative samples on ImageNet $256\times256$, conditioned on the indicated classes, generated with JiT-H/16 + Dual Head and CFG of 2.8.}
  \label{fig:supp:qualitative-4}
\end{figure}

\begin{figure}[h]
    \centering
    \includegraphics[width=0.97\linewidth]{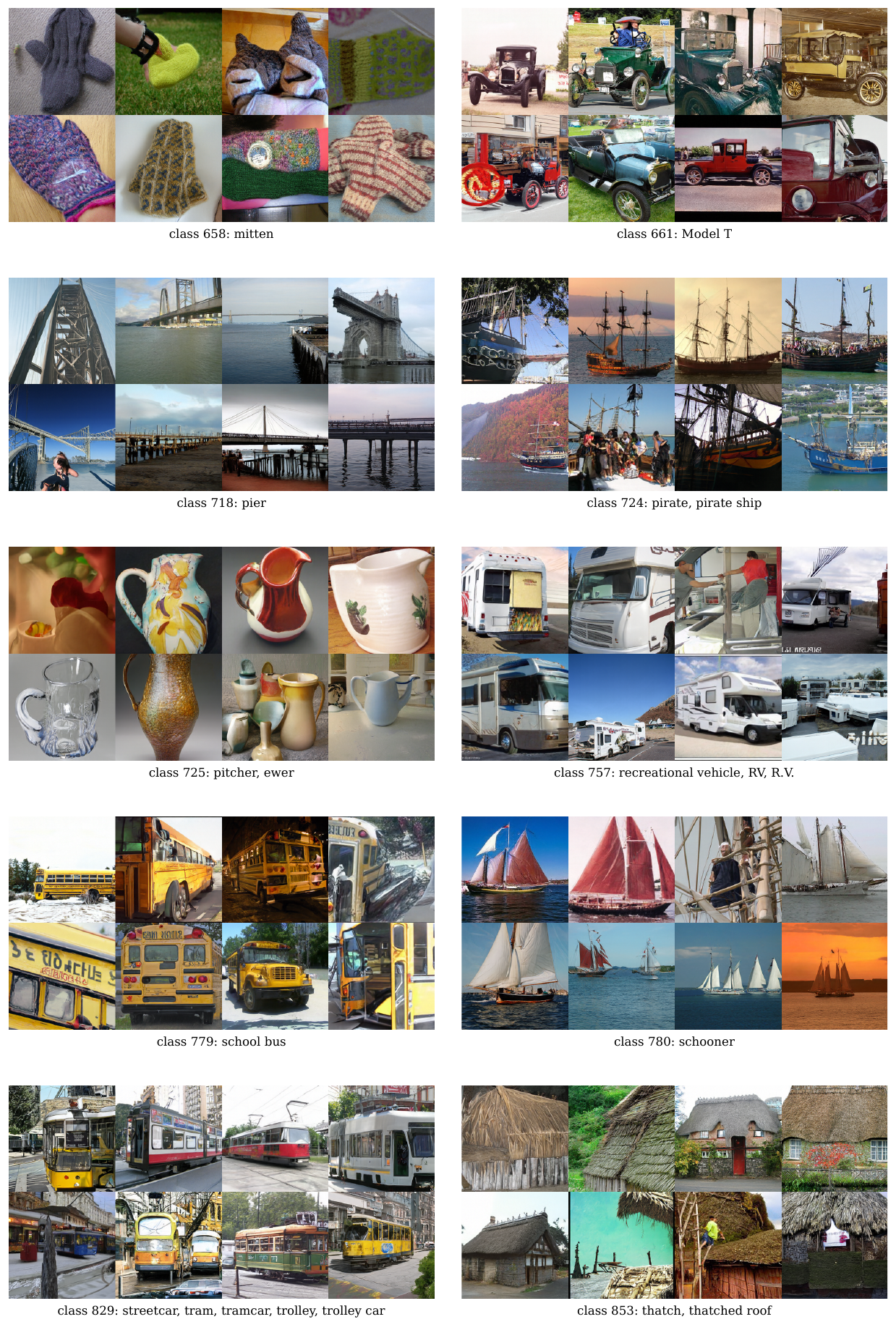}
  \caption{Qualitative samples on ImageNet $256\times256$, conditioned on the indicated classes, generated with JiT-L/16 + Dual Head and CFG of 2.6.}
  \label{fig:supp:qualitative-5}
\end{figure}

\end{document}

%% file: math_commands.tex
\usepackage{amsmath,amsfonts,bm}

\def\eqref#1{equation~\ref{#1}}

\def\1{\bm{1}}

\DeclareMathAlphabet{\mathsfit}{\encodingdefault}{\sfdefault}{m}{sl}
\SetMathAlphabet{\mathsfit}{bold}{\encodingdefault}{\sfdefault}{bx}{n}

\newcommand{\R}{\mathbb{R}}

%% file: commands.tex
\newcommand{\epsclip}{\ensuremath{\tau_{\mathrm{clip}}}\xspace}  %
\newcommand{\x}{\ensuremath{\mathbf{x}}\xspace}

\newcommand{\xpred}{\hat{\mathbf{x}}_{\theta}}
\newcommand{\vpred}{\hat{\mathbf{v}}_{\theta}}

\newcommand{\xlin}{\hat{\mathbf{x}}_{\mathrm{lin}}\xspace}
\newcommand{\xconv}{\hat{\mathbf{x}}_{\mathrm{conv}}\xspace}

%% file: figures/blurriness/blurriness.tex
\noindent
\begin{tikzpicture}[
  title/.style={
    font=\fontsize{15}{15}\selectfont,
    anchor=south
  }
]

\def\imagewidth{4cm}
\def\pairgap{0.2cm}

\node[anchor=south west, inner sep=0pt] (img0) at (0,0)
  {\includegraphics[
    width=\imagewidth,
    height=\imagewidth
  ]{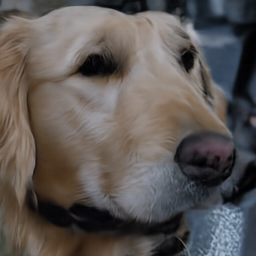}};

\node[anchor=south west, inner sep=0pt] (img1) at
  (img0.south east)
  {\includegraphics[
    width=\imagewidth,
    height=\imagewidth,
    trim=0 2.8cm 2.8cm 0,
    clip
  ]{figures/blurriness/blurry.png}};

\node[anchor=south west, inner sep=0pt] (img2) at
  ($(img1.south east)+(\pairgap,0)$)
  {\includegraphics[
    width=\imagewidth,
    height=\imagewidth,
    trim=0 2.8cm 2.8cm 0,
    clip
  ]{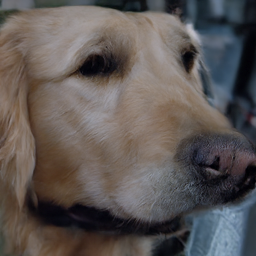}};

\node[anchor=south west, inner sep=0pt] (img3) at
  (img2.south east)
  {\includegraphics[
    width=\imagewidth,
    height=\imagewidth,
  ]{figures/blurriness/blurry_conv.png}};

\node[title, yshift=-3pt] at
  ($(img0.north)!0.5!(img1.north)$)
  {JiT (sharpness 91.0)};

\node[title, yshift=-3pt] at
  ($(img2.north)!0.5!(img3.north)$)
  {JiT with conv layers (sharpness 154.0)};

\end{tikzpicture}

%% file: figures/artifacts/artifacts.tex
\begin{tikzpicture}[
  title/.style={
    font=\fontsize{21}{21}\selectfont,
    anchor=south
  },
  patchlabel/.style={
    text=white,
    font=\fontsize{21}{21}\selectfont,
    align=center
  }
]

\def\imagewidth{6cm}
\pgfmathsetlengthmacro{\patchside}{\imagewidth/8}

\node[anchor=south west, inner sep=0] (img0) at (0,0)
  {\includegraphics[width=\imagewidth]{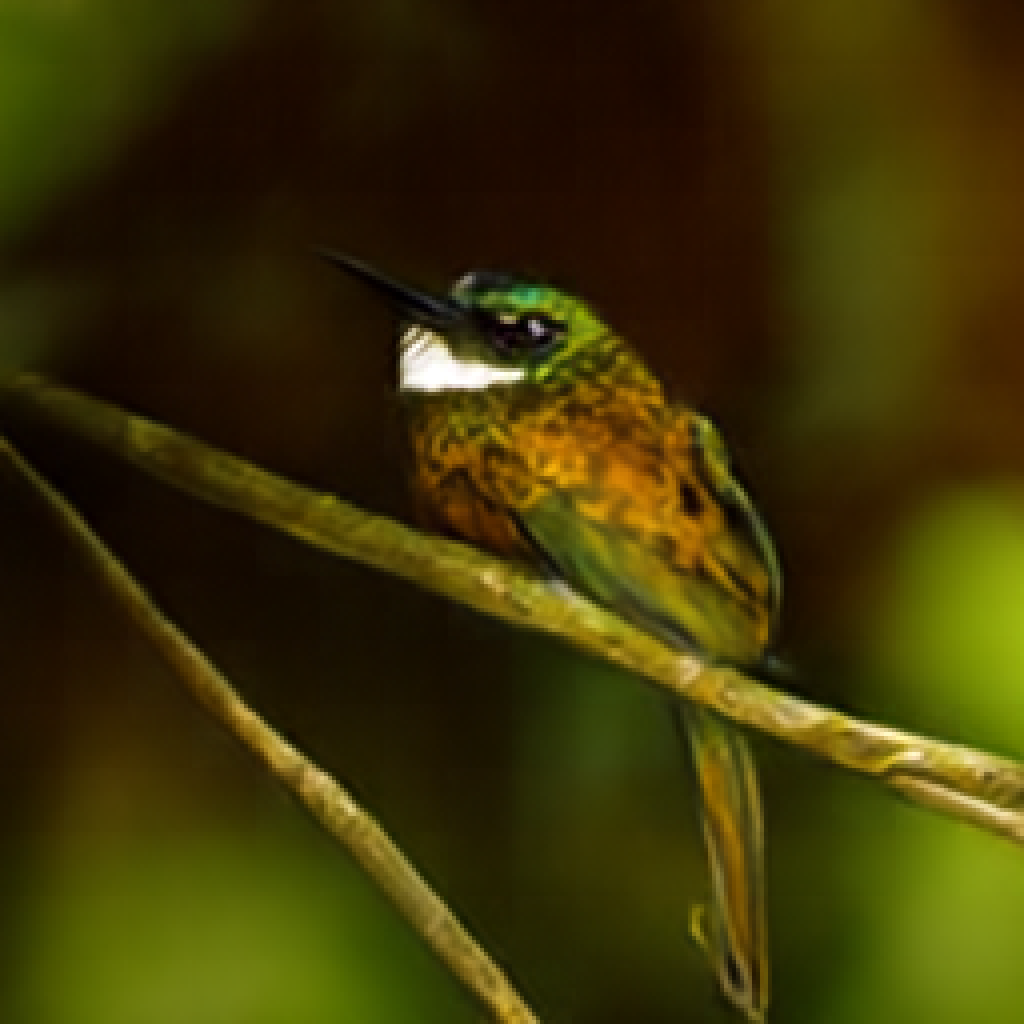}};

\node[anchor=south west, inner sep=0] (img1) at ($(img0.south east)$)
  {\includegraphics[width=\imagewidth]{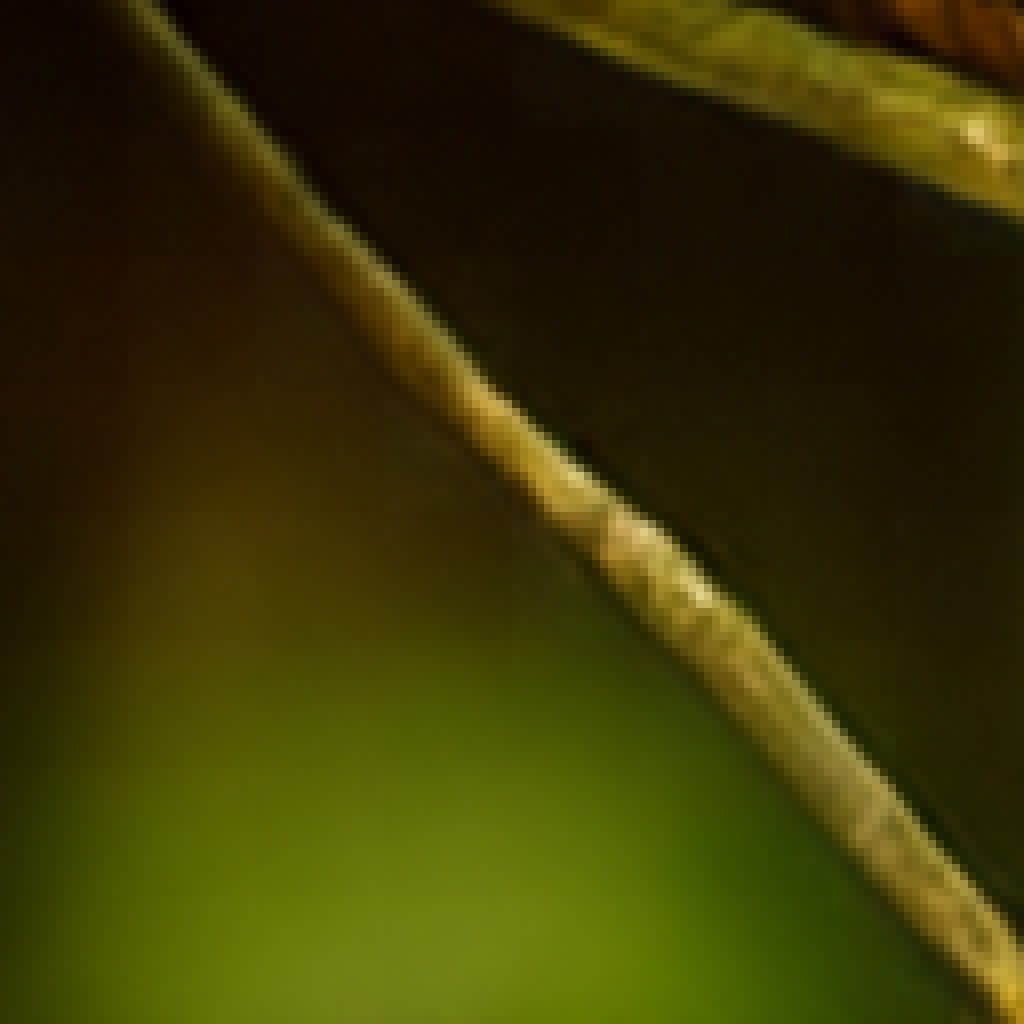}};

\node[anchor=south west, inner sep=0] (img2) at ($(img1.south east)$)
  {\includegraphics[width=\imagewidth]{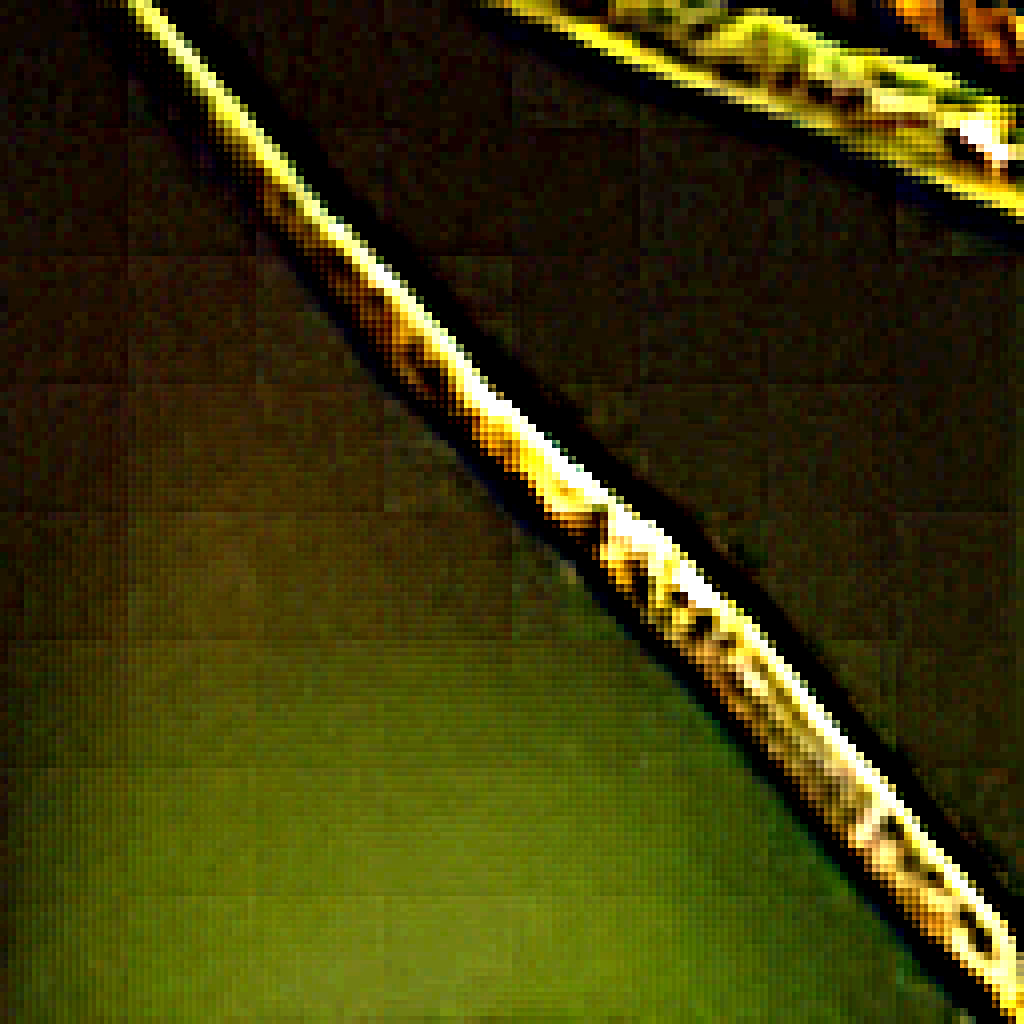}};

\node[title, yshift=-2pt] at (img0.north) {Original 256x256};
\node[title, yshift=2pt] at (img1.north) {Cut 128x128};
\node[title, yshift=-2pt] at (img2.north) {Sharpened 128x128};

\def\patchx{0.003}
\def\patchy{0.003}

\coordinate (patchSW) at
  ($(img2.south west)! \patchx ! (img2.south east)
    + (0,\patchy*\imagewidth)$);

\draw[
  white,
  line width=1.5pt
]
  (patchSW) rectangle ++(\patchside,\patchside);

\node[
  patchlabel,
  font=\fontsize{21}{21}\selectfont,
  anchor=south west
] at ($(img2.south west) + (0.10*\imagewidth,0.16*\imagewidth)$)
{Patch size};

\node[
  patchlabel,
  font=\fontsize{21}{21}\selectfont,
  anchor=south west
] at ($(img2.south west) + (0.20*\imagewidth,0.035*\imagewidth)$)
{16x16};

\end{tikzpicture}

%% file: figures/jit_diagram.tex
\usetikzlibrary{arrows.meta,positioning,calc,decorations.pathreplacing}%

\begingroup%

\definecolor{cInk}{HTML}{1B2430}%
\definecolor{cRule}{HTML}{5A6472}%
\definecolor{cTok}{HTML}{E9EEF4}%
\definecolor{cTokB}{HTML}{E3E9F7}%
\definecolor{cTokS}{HTML}{8894A3}%
\definecolor{cBlue}{HTML}{2B6CB0}%
\definecolor{cBlueF}{HTML}{DBE8F7}%
\definecolor{cTeal}{HTML}{2C7A7B}%
\definecolor{cTealF}{HTML}{D5EBE8}%
\definecolor{cViol}{HTML}{6B46C1}%
\definecolor{cViolF}{HTML}{E7E1F5}%
\definecolor{cAcc}{HTML}{C05621}%
\definecolor{cSky}{HTML}{BFDCEF}%
\definecolor{cSun}{HTML}{F2C14E}%
\definecolor{cHillA}{HTML}{7FA6A0}%
\definecolor{cHillB}{HTML}{4E7C74}%
\definecolor{cGrass}{HTML}{A9C489}%

\tikzset{%
  every node/.style={font=\sffamily\small, text=cInk, inner sep=1.5pt},%
  flow/.style={draw=cRule, line width=0.9pt,%
               -{Stealth[length=4.6pt,width=3.6pt]}},%
  sflow/.style={draw=cRule, line width=0.7pt,%
                -{Stealth[length=4pt,width=3.2pt]}},%
  block/.style={line width=0.9pt, rounded corners=2.5pt, inner sep=0pt},%
  ccap/.style={font=\sffamily\scriptsize, text=cRule},%
  lead/.style={draw=cRule, line width=0.5pt, opacity=0.55,%
               dash pattern=on 1.8pt off 1.8pt},%
  brc/.style={draw=cRule, line width=0.6pt, decorate,%
              decoration={brace, amplitude=4pt, raise=1pt}},%
}%

\newcommand{\vitphoto}{%
  \begin{scope}%
    \clip (0,0) rectangle (2.8,2.8);%
    \fill[cSky] (0,0) rectangle (2.8,2.8);%
    \fill[white, opacity=0.9] (0.28,2.53) circle (0.21);%
    \fill[white, opacity=0.9] (0.55,2.46) circle (0.15);%
    \fill[white, opacity=0.9] (0.08,2.44) circle (0.13);%
    \fill[cSun] (2.16,2.22) circle (0.28);%
    \fill[cHillA] (1.15,0) -- (2.25,1.30) -- (3.30,0) -- cycle;%
    \fill[cHillB] (-0.25,0) -- (1.05,1.72) -- (2.30,0) -- cycle;%
    \fill[cGrass] (0,0) rectangle (2.8,0.52);%
  \end{scope}%
}%
\newcommand{\vitgrid}[1]{%
  \foreach \i in {1,...,5}{%
    \draw[white, line width=#1, opacity=0.85] (\i*2.8/6,0) -- (\i*2.8/6,2.8);%
    \draw[white, line width=#1, opacity=0.85] (0,\i*2.8/6) -- (2.8,\i*2.8/6);}%
  \draw[draw=cRule, line width=0.9pt] (0,0) rectangle (2.8,2.8);%
}%
\newcommand{\vittokens}[3]{%
  \foreach \yy in {2.60, 2.12, 1.64, 0.68, 0.20}{%
    \draw[fill=#2, draw=#3, line width=0.6pt, rounded corners=1pt]%
      (#1,\yy-0.17) rectangle (#1+1.2,\yy+0.17);%
    \foreach \k in {1,...,5}{%
      \draw[white, line width=0.5pt]%
        (#1+\k*0.2,\yy-0.17) -- (#1+\k*0.2,\yy+0.17);}}%
  \node[text=cRule] at (#1+0.6,1.14) {$\vdots$};%
}%

\begin{tikzpicture}

\vitphoto
\vitgrid{0.7pt}
\draw[draw=cAcc, line width=1.1pt] (0,2.8-2.8/6) rectangle (2.8/6,2.8);
\node[ccap, anchor=north] at (1.4,-0.28) {$H\times W\times 3$ image};

\draw[lead] (0,2.8)     -- (0,3.02);
\draw[lead] (2.8/6,2.8) -- (1.4,3.02);
\begin{scope}
  \clip (0,3.02) rectangle (1.4,4.42);
  \begin{scope}[shift={(0,3.02)}, scale=3, shift={(0,-2.33333)}]
    \vitphoto
  \end{scope}
\end{scope}
\foreach \k in {1,...,15}{%
  \draw[white, line width=0.15pt, opacity=0.55]
    (\k*1.40/16,3.02) -- (\k*1.40/16,4.42);
  \draw[white, line width=0.15pt, opacity=0.55]
    (0,3.02+\k*1.40/16) -- (1.4,3.02+\k*1.40/16);}
\draw[draw=cAcc, line width=1.1pt] (0,3.02) rectangle (1.4,4.42);
\node[ccap, anchor=west, align=left] at (1.55,3.72)
     {one patch:\\[1pt] $16\times16\times3$\\[1pt] $=768$ values};

\draw[flow] (2.95,1.4) -- (4.53,1.4);
\node[ccap, align=center, anchor=south] at (3.74,1.52) {patchify\\+ flatten};

\vittokens{4.55}{cTok}{cTokS}
\node[ccap, anchor=north, align=center] at (4.5,-0.28)
     {$N=HW/16^{2}$\\ tokens, $N\times 768$};

\draw[flow] (5.9,1.4) -- (6.28,1.4);

\draw[block, fill=cBlueF, draw=cBlue] (6.3,-0.10) rectangle (7.25,2.90);
\node[rotate=90, text=cBlue] at (6.775,1.4) {Linear $768\rightarrow 128$};
\draw[flow] (7.25,1.4) -- (7.53,1.4);
\draw[block, fill=cBlueF, draw=cBlue] (7.55,-0.10) rectangle (8.5,2.90);
\node[rotate=90, text=cBlue] at (8.025,1.4) {Linear $128\rightarrow 768$};

\draw[brc] (6.3,2.98) -- (8.5,2.98);
\node[anchor=south, align=center] at (7.4,3.20)
     {bottleneck\\ patch embedding};
\node[ccap, anchor=north] at (7.4,-0.28) {$768\rightarrow 128\rightarrow 768$};

\draw[block, fill=cTealF, draw=cTeal, opacity=0.55]
      (9.14,0.04) rectangle (10.84,3.04);
\draw[block, fill=cTealF, draw=cTeal, opacity=0.75]
      (9.07,-0.03) rectangle (10.77,2.97);
\draw[block, fill=cTealF, draw=cTeal] (9,-0.10) rectangle (10.7,2.90);
\node[rotate=90, text=cTeal, align=center] at (9.85,1.4)
     {Transformer\\ encoder $\times L$};
\draw[flow] (8.52,1.4) -- (8.98,1.4);
\node[ccap, anchor=north] at (9.85,-0.28) {$N\times 768$};
\draw[flow] (10.86,1.4) -- (11.13,1.4);

\draw[block, fill=cViolF, draw=cViol] (11.15,-0.10) rectangle (12.1,2.90);
\node[rotate=90, text=cViol] at (11.625,1.4) {Linear $768\rightarrow 768$};
\node[ccap, anchor=south] at (11.625,2.98) {patch head};
\draw[flow] (12.12,1.4) -- (12.55,1.4);

\vittokens{12.60}{cTokB}{cTokS}
\node[ccap, anchor=north, align=center] at (13.2,-0.28)
     {$N\times 768$\\ $=N\times(16\!\times\!16\!\times\!3)$};

\draw[flow] (13.95,1.4) -- (15.92,1.4);
\node[ccap, align=center, anchor=south] at (14.94,1.52) {reshape\\+ unpatchify};

\begin{scope}[shift={(16,0)}]
  \vitphoto
  \fill[white, opacity=0.10] (0,0) rectangle (2.8,2.8);
  \vitgrid{1.1pt}
  \node[ccap, anchor=north, align=center] at (1.4,-0.28)
       {predicted $16\times16$ patches\\ $H\times W\times 3$};
\end{scope}

\end{tikzpicture}%
\endgroup%

%% file: figures/convjit_diagram.tex
\usetikzlibrary{arrows.meta,positioning,calc}
\begingroup
\definecolor{cInk}{HTML}{1B2430}
\definecolor{cRule}{HTML}{5A6472}
\definecolor{cTeal}{HTML}{2C7A7B}
\definecolor{cTealF}{HTML}{D5EBE8}
\definecolor{cViol}{HTML}{6B46C1}
\definecolor{cViolF}{HTML}{E7E1F5}

\begin{tikzpicture}[
  every node/.style={
    font=\sffamily\normalsize, text=cInk, inner sep=2pt
  },
  flow/.style={
    draw=cRule, line width=0.9pt,
    -{Stealth[length=5pt,width=4pt]}
  },
  block/.style={
    line width=0.9pt, rounded corners=2.5pt,
    minimum width=3.4cm, minimum height=1.65cm,
    align=center
  },
  ccap/.style={font=\sffamily\small, text=cRule}
]
\node (input) at (0,0) {$x_t$};

\node[block, draw=cTeal, fill=cTealF] (backbone) at (3,0) {};
\node[text=cTeal] at ($(backbone.center)+(0,0.46)$)
  {JiT transformer};
\node[ccap] at (backbone.center) {big patches};
\node[ccap] at ($(backbone.center)+(0,-0.46)$)
  {$h=b(x_t,t,c)$};

\node[block, draw=cViol, fill=cViolF] (head) at (7.6,0) {};
\node[text=cViol] at ($(head.center)+(0,0.46)$)
  {pixel head};
\node[ccap] at (head.center) {small patches};
\node[ccap] at ($(head.center)+(0,-0.46)$)
  {$x_0=g(h,x_t,t,c)$};

\node (output) at (10.6,0) {$x_0$};

\draw[flow] (input.east) -- (backbone.west);
\draw[flow] (backbone.east) --
  node[above] {$h$} (head.west);
\draw[flow] (head.east) -- (output.west);

\node[ccap, above=5pt of input, align=center] {noisy\\image};
\node[ccap, above=5pt of output, align=center] {predicted\\clean image};

\coordinate (branch) at (0.75,0);
\fill[cRule] (branch) circle (1.3pt);
\draw[flow] (branch) -- (0.75,-1.35)
  -- node[above, ccap] {$x_t$} (7.6,-1.35)
  -- (head.south);
\end{tikzpicture}
\endgroup

%% file: figures/block_diagram.tex
\usetikzlibrary{arrows.meta,positioning,calc}
\begingroup

\definecolor{cInk}{HTML}{1B2430}
\definecolor{cRule}{HTML}{5A6472}
\definecolor{cBox}{HTML}{EDF1F6}
\definecolor{cBlue}{HTML}{2B6CB0}
\definecolor{cBlueF}{HTML}{DBE8F7}

\begin{tikzpicture}[
  every node/.style={
    font=\sffamily\normalsize,
    text=cInk,
    inner sep=2pt
  },
  flow/.style={
    draw=cRule,
    line width=0.9pt,
    -{Stealth[length=5pt,width=4pt]}
  },
  cond/.style={
    flow,
    draw=cBlue,
    dash pattern=on 3pt off 2pt
  },
  block/.style={
    draw=cRule,
    fill=cBox,
    line width=0.9pt,
    rounded corners=2.5pt,
    minimum height=0.85cm,
    text height=0.24cm,
    text depth=0.06cm,
    align=center,
    inner sep=4pt,
    outer sep=0pt
  },
  op/.style={
    circle,
    draw=cRule,
    fill=white,
    line width=0.9pt,
    minimum size=0.45cm,
    inner sep=0pt,
    outer sep=0pt
  },
  ccap/.style={
    font=\sffamily\small,
    text=cBlue,
    inner sep=0pt,
    anchor=south
  }
]

\node (x) at (-0.4,0) {$x$};

\node[block, text width=1.9cm]
  (conv) at (1.65,0) {$3\times3$ Conv};

\node[block, text width=2.0cm]
  (norm) at (4.15,0) {GroupNorm};

\node[block, text width=1.0cm]
  (mlp) at (6.2,0) {MLP};

\node[op, draw=cBlue, fill=cBlueF]
  (gate) at (7.65,0) {$\odot$};

\node[op] (add) at (8.55,0) {$+$};
\node (y) at (9.6,0) {$y$};

\draw[flow] (x) -- (conv);
\draw[flow] (conv) -- (norm);
\draw[flow] (norm) -- (mlp);
\draw[flow] (mlp) -- (gate);
\draw[flow] (gate) -- (add);
\draw[flow] (add) -- (y);

\node[
  block,
  draw=cBlue,
  fill=cBlueF,
  text=cBlue,
  text width=3.2cm,
  minimum height=0.7cm
] (condition) at (5.9,1.55)
  {SiLU -- $1\times1$ Conv};

\node[left=0.35cm of condition] (c) {$c$};
\draw[flow] (c) -- (condition);

\coordinate (split) at (5.9,0.7);

\draw[draw=cBlue, line width=0.9pt,
      dash pattern=on 3pt off 2pt]
  (condition.south) -- (split);

\fill[cBlue] (split) circle (1.2pt);

\draw[cond] (split) -- (4.15,0.7) -- (norm.north);
\draw[cond] (split) -- (7.65,0.7) -- (gate.north);

\node[ccap] at (4.25,0.9) {shift, scale};
\node[ccap] at (7.05,0.9) {gate};

\coordinate (branch) at (0.05,0);
\fill[cRule] (branch) circle (1.2pt);

\draw[flow] (branch)
  -- (0.05,-0.8)
  -- (8.55,-0.8)
  -- (add.south);

\end{tikzpicture}
\endgroup